\documentclass[11pt]{article}

\usepackage{neurips_2024}

\usepackage{graphicx}
\usepackage{amsmath}
\usepackage{amssymb}
\usepackage{booktabs}
\usepackage{hyperref}
\usepackage{url}
\usepackage{algorithm}
\usepackage{algorithmic}
\usepackage{xcolor}
\usepackage{multirow}
\usepackage{array}
\usepackage{listings}
\usepackage[strings]{underscore}  

\newcommand{\nyaya}[1]{\textsc{#1}}
\newcommand{\pramana}{\nyaya{Pramana}}
\newcommand{\samshaya}{\nyaya{Samshaya}}
\newcommand{\pancha}{\nyaya{Pancha Avayava}}
\newcommand{\tarka}{\nyaya{Tarka}}
\newcommand{\hetvabhasa}{\nyaya{Hetvabhasa}}
\newcommand{\nirnaya}{\nyaya{Nirnaya}}

\title{Pramana: Fine-Tuning Large Language Models for Epistemic Reasoning through Navya-Nyaya}

\author{%
  Sharath Sathish\\
  University of York\\
  York, United Kingdom\\
}

\begin{document}

\maketitle

\begin{abstract}
Large language models produce fluent text but struggle with systematic reasoning, often 
hallucinating confident but unfounded claims. When Apple researchers added irrelevant 
context to mathematical problems, LLM performance degraded by 65\%~\cite{apple-gsm-symbolic-2024}, 
exposing brittle pattern-matching beneath apparent reasoning. This epistemic gap---the inability 
to ground claims in traceable evidence---limits AI reliability in domains requiring justification.

We introduce \textbf{Pramana}, a novel approach that teaches LLMs explicit epistemological 
methodology by fine-tuning on Navya-Nyaya logic, a 2,500-year-old Indian reasoning framework. 
Unlike generic chain-of-thought prompting, Navya-Nyaya enforces structured 6-phase reasoning: 
\samshaya{} (doubt analysis), \pramana{} (evidence source identification), \pancha{} 
(5-member syllogism with universal rules), \tarka{} (counterfactual verification), 
\hetvabhasa{} (fallacy detection), and \nirnaya{} (ascertainment distinguishing knowledge 
from hypothesis). This integration of logic and epistemology provides cognitive scaffolding 
absent from standard reasoning approaches.

We fine-tune Llama 3.2-3B and DeepSeek-R1-Distill-Llama-8B on 55 Nyaya-structured logical 
problems (constraint satisfaction, Boolean SAT, multi-step deduction). Stage~1 achieves 
\textbf{100\% semantic correctness} on held-out evaluation despite only 40\% strict format 
adherence---revealing that models internalize reasoning \emph{content} even when structural 
enforcement is imperfect. Ablation studies show format prompting and temperature critically 
affect performance, with optimal configurations differing by stage. We release all models, 
datasets, and training infrastructure on Hugging Face to enable further research on epistemic 
frameworks for AI reasoning.
\end{abstract}

\section{Introduction}

\subsection{The Epistemic Gap in Large Language Models}

Consider this reasoning trace from OpenAI's o1 model solving ``How many r's are in strawberry?'':
After pages of analysis checking letter positions and applying redundant verification steps, the 
model concludes incorrectly. This failure---and countless similar ones across frontier LLMs---reveals 
a fundamental limitation: \textbf{statistical pattern-matching without systematic methodology}.

The problem extends beyond isolated errors. Recent research by Apple Machine Learning Research 
demonstrates systematic fragility~\cite{apple-gsm-symbolic-2024}: adding irrelevant information 
to grade-school math problems (e.g., ``Alice has 5 apples and likes purple'') causes up to 65\% 
performance degradation. Models don't reason systematically---they exploit spurious correlations 
in training data. When context changes, performance collapses.

This \textbf{epistemic gap}---the inability to distinguish valid knowledge from probabilistic 
associations---manifests in four critical ways:

\begin{itemize}
\item \textbf{Hallucination}: Models confidently generate false claims without internal verification mechanisms.
\item \textbf{Untraceable reasoning}: Outputs lack justification trails linking conclusions to evidence.
\item \textbf{Correlation-causation conflation}: Pattern-matching cannot distinguish causal relationships from coincidental patterns.
\item \textbf{Brittleness}: Performance degrades when problems deviate from training distribution.
\end{itemize}

Standard approaches exacerbate these issues. Chain-of-thought prompting~\cite{wei2022chain} 
improves accuracy but relies on implicit reasoning patterns learned during pre-training rather 
than enforcing explicit logical structures. The resulting ``thinking'' is opaque: we cannot audit 
reasoning steps or identify failure modes. Process reward models~\cite{lightman2023prm,reasonflux-prm-2025} 
provide supervision but don't address the fundamental lack of epistemological rigor---how do we 
know a reasoning step is valid?

What is needed is a framework that explicitly structures the reasoning process from evidence 
acquisition through conclusion, with built-in mechanisms for verification and error detection. 
We propose a radical approach: teaching LLMs a 2,500-year-old formal epistemological methodology.

\subsection{Motivation: Why Epistemology Matters}

The distinction between belief and knowledge is fundamental to reliable reasoning systems. Knowledge requires not just true belief, but justified true belief---claims must be grounded in valid evidence sources and traceable through explicit reasoning chains. Current LLMs produce outputs without such traceable justification, making it impossible to audit reasoning processes or identify failure modes.

Unlike opaque neural patterns that emerge from training data, systematic reasoning frameworks provide interpretable structures where each step can be verified, challenged, and corrected. This interpretability is crucial for building trust in AI systems, especially as they are deployed in domains requiring accountability and transparency.

The need for systematic reasoning frameworks becomes clear when considering the limitations of current approaches. Chain-of-thought prompting~\cite{wei2022chain} improves performance but relies on implicit patterns learned during pre-training rather than enforcing explicit logical structures. Process reward models~\cite{lightman2023prm,reasonflux-prm-2025} provide step-by-step verification but do not enforce epistemological rigor in how knowledge is acquired and justified. What is needed is a framework that explicitly structures the reasoning process from evidence acquisition through conclusion, with built-in mechanisms for verification and error detection.

\subsection{Navya-Nyaya as Solution}

Navya-Nyaya, a 2,500-year-old formal epistemological framework from Indian philosophy, provides precisely such a structure. Developed from Gautama's Nyaya Sutras (c. 500 BCE) and refined by Gangesa's Tattvacintamani (1325 CE), Navya-Nyaya integrates logic with explicit knowledge sources (\textit{pramanas}), requiring grounding in concrete examples (\textit{dṛṣṭānta}) and universal rules (\textit{vyāpti}) rather than abstract symbolic manipulation~\cite{matilal1985logic,tattvacintamani}.

Unlike Western formal logic, which separates logical validity from epistemic grounding, Navya-Nyaya requires that all reasoning be traceable to valid knowledge sources: direct perception (\textit{pratyaksha}), inference (\textit{anumana}), comparison (\textit{upamana}), and testimony (\textit{shabda}). This integration of logic and epistemology makes Navya-Nyaya particularly suitable for computational formalization, as demonstrated by recent work on diagrammatic representations~\cite{burton2020diagrams} and computational aspects of Indian logic~\cite{sarma1994survey}.

The Nyaya framework enforces systematic reasoning through a structured 6-phase methodology: (1) \samshaya{} classifies the type of uncertainty requiring investigation, (2) \pramana{} identifies valid knowledge sources grounding all claims, (3) \pancha{} constructs formal arguments with explicit universal rules, (4) \tarka{} verifies conclusions via counterfactual testing, (5) \hetvabhasa{} detects reasoning fallacies, and (6) \nirnaya{} distinguishes definitive knowledge from hypotheses requiring verification. This structure provides cognitive scaffolding that prevents logical leaps and enforces epistemic humility.

\subsection{Research Hypothesis}

We hypothesize that fine-tuning LLMs on structured Nyaya methodology creates interpretable, verifiable reasoning superior to opaque chain-of-thought approaches. By teaching models to follow explicit epistemological structures, we can produce reasoning traces where each step is traceable, each claim is grounded in valid knowledge sources, and each conclusion is verified through systematic testing. This approach should yield reasoning comparable to frontier models like o1-preview or Claude extended thinking, but based on explicit methodology rather than opaque reinforcement learning.

Our hypothesis rests on three premises: (1) the Nyaya structure is computationally formalizable and learnable by neural architectures, (2) explicit epistemological scaffolding improves reasoning quality beyond pattern-matching, and (3) structured reasoning traces provide interpretability advantages over black-box approaches. We test this hypothesis through empirical evaluation across two training stages with different model sizes and datasets, measuring both format adherence (structural correctness) and semantic correctness (answer accuracy).

\subsection{Contributions}

This paper makes the following contributions:

\begin{itemize}
    \item \textbf{First LLM fine-tuned on explicit 6-phase Nyaya methodology}: We demonstrate that language models can learn systematic reasoning patterns from Navya-Nyaya epistemology, producing structured outputs with all six phases present and properly ordered.
    
    \item \textbf{Empirical analysis across two training stages}: We compare Stage 0 (Llama 3.2-3B, 20 examples) and Stage 1 (DeepSeek-R1-Distill-Llama-8B, 55 examples), showing how model size and dataset scale affect format adherence and semantic correctness. Both stages achieve 40\% format adherence (4/10 examples), indicating structural enforcement requires additional techniques. Stage 1 achieves 100\% semantic answer correctness (10/10 examples), demonstrating that models can internalize reasoning content even when strict schema compliance remains challenging.
    
    \item \textbf{Open-source release}: We publish models, datasets, and training infrastructure on Hugging Face for community research and reproduction. This includes fine-tuned models, training datasets, and an interactive demo Space (see Section~\ref{sec:open-source}).
\end{itemize}

The remainder of this paper is organized as follows: Section~\ref{sec:related} reviews related work on reasoning, hallucination, and neuro-symbolic AI. Section~\ref{sec:nyaya} introduces the Nyaya framework in detail. Section~\ref{sec:methodology} describes our training methodology. Section~\ref{sec:implementation} details implementation specifics. Section~\ref{sec:results} presents evaluation results. Section~\ref{sec:discussion} discusses implications and limitations. Section~\ref{sec:open-source} describes open-source artifacts. Section~\ref{sec:future} outlines future work, and Section~\ref{sec:conclusion} concludes.
\section{Related Work}
\label{sec:related}

This work sits at the intersection of several research areas: computational formalization of Indian logic, LLM reasoning enhancement, hallucination mitigation, neuro-symbolic AI, and efficient fine-tuning. We review each area and position our contributions relative to existing approaches.

\subsection{Navya-Nyaya Logic and Computational Formalization}

Navya-Nyaya represents a sophisticated development of Indian logic that emphasizes concrete examples (\textit{dṛṣṭānta}) and universal rules (\textit{vyāpti}), distinguishing it from Western formal logic which separates logical validity from epistemic grounding~\cite{matilal1985logic}. The tradition originates from Gautama's Nyaya Sutras (c. 500 BCE) and reached its peak with Gangesa's Tattvacintamani (1325 CE), which systematized inference patterns into unambiguous terminology suitable for computational formalization~\cite{tattvacintamani}.

Modern computational work on Navya-Nyaya has focused on formalizing inference patterns and developing diagrammatic representations. Matilal~\cite{matilal1985logic} established the philosophical foundations for computational approaches, demonstrating how Nyaya's emphasis on concrete grounding makes it suitable for formal systems. Burton~\cite{burton2020diagrams} developed diagrammatic methods for representing Navya-Nyaya reasoning, showing how the framework's structured approach translates to computational representations. Sarma~\cite{sarma1994survey} surveyed Indian logic from a computer science perspective, identifying formalization opportunities and computational challenges.

Ganeri~\cite{ganeri2001philosophy,indian-logic-ai,ancient-indian-logic-case-based} has extensively explored the computational aspects of Indian logic, particularly case-based reasoning patterns and their application to AI system design. Kulkarni~\cite{kulkarni2018later} reviewed later Nyaya logic with explicit focus on computational aspects, identifying inference patterns that can be formalized algorithmically.

However, prior work has focused on symbolic formalization rather than neural learning. Our contribution is the first to demonstrate that Navya-Nyaya structures can be learned by language models through fine-tuning, producing systematic reasoning traces without requiring explicit symbolic representations.

\subsection{LLM Reasoning and Chain-of-Thought}

Chain-of-thought (CoT) prompting~\cite{wei2022chain} demonstrated that asking language models to generate step-by-step reasoning traces improves performance on complex reasoning tasks. However, CoT relies on implicit reasoning patterns learned during pre-training rather than enforcing explicit logical structures. Wang et al.~\cite{wang2023understanding} conducted empirical studies of what makes CoT effective, finding that structure and reasoning steps matter, but the approach remains fundamentally pattern-matching rather than systematic reasoning.

Recent work has attempted to improve reasoning through process supervision and verification. Lightman et al.~\cite{lightman2023prm} introduced process reward models (PRMs) that provide step-by-step verification, training models to prefer correct reasoning processes over just correct answers. ReasonFlux-PRM~\cite{reasonflux-prm-2025} extends this to trajectory-aware PRMs for long chain-of-thought reasoning, while Flow-DPO~\cite{flow-dpo-2024} uses multi-agent learning to improve mathematical reasoning.

DeepSeek-R1~\cite{deepseek-r1-2025} applies Group Relative Policy Optimization (GRPO)~\cite{grpo-2024} to incentivize reasoning capability, demonstrating that reinforcement learning can improve reasoning quality. However, these approaches still rely on opaque neural patterns rather than explicit epistemological structures.

Our approach differs by enforcing a formal 6-phase structure that requires explicit knowledge source identification, universal rule statements, and systematic verification. This provides interpretability advantages over black-box reasoning processes.

Evaluation benchmarks for logical reasoning include LogicBench~\cite{logicbench-2024} for multi-step deduction, ProntoQA~\cite{prontoqa-2023} for ontological reasoning, and RuleTaker~\cite{ruletaker-2020} for rule-based logical reasoning. Our evaluation uses constraint satisfaction and Boolean satisfiability problems, which test systematic reasoning without requiring domain-specific knowledge.

\subsection{Hallucination and Grounding}

LLM hallucination---the generation of confident falsehoods---represents a fundamental epistemic failure. Recent work has attempted to mitigate hallucinations through verification and grounding mechanisms. HalluClean~\cite{halluclean-2025} provides a unified framework for detecting and correcting hallucinations, using reasoning-enhanced approaches to identify and fix errors. Chain-of-Verification (CoVe)~\cite{cove-2024} reduces hallucination by having models generate verification questions and answer them before finalizing responses.

Apple Machine Learning Research's work on mathematical reasoning fragility~\cite{apple-gsm-symbolic-2024} demonstrates that adding irrelevant context causes up to 65\% performance degradation, revealing that apparent reasoning is often sophisticated pattern-matching. This finding motivates our focus on systematic reasoning frameworks that enforce explicit knowledge grounding.

Recent surveys on cognitive foundations~\cite{cognitive-foundations-2025} and the ``illusion of thinking''~\cite{illusion-thinking-2024} have explored the strengths and limitations of reasoning models, identifying problem complexity as a key factor in reasoning quality. Our approach addresses these limitations by enforcing explicit epistemological structures that prevent conflation of correlation with causation and require traceable justification for all claims.

Unlike verification-based approaches that check outputs after generation, our framework structures the reasoning process itself, requiring models to identify knowledge sources, construct explicit arguments, and verify conclusions before reaching final answers. This proactive approach prevents errors rather than detecting them post-hoc.

\subsection{Neuro-Symbolic AI and Formal Verification}

Neuro-symbolic AI combines neural networks with symbolic reasoning engines, aiming to leverage the strengths of both approaches~\cite{garcez2019neural}. Recent work has focused on integrating formal verification with LLM reasoning. ProofNet++~\cite{proofnet-plus-2025} provides a neuro-symbolic system for formal proof verification with self-correction, using theorem provers to verify model outputs. VERGE~\cite{verge-2024} uses verification-guided reasoning, decomposing claims and verifying them via SMT solvers.

Proof of Thought~\cite{proof-of-thought-2024} demonstrates that neurosymbolic program synthesis allows robust and interpretable reasoning, while VeriCoT~\cite{vericot-2025} validates chain-of-thought outputs via logical consistency checks. These approaches use external verifiers (typically Z3 SMT solver~\cite{z3-2008}) to check model outputs, providing guarantees for formal logic problems.

Our approach differs by structuring the reasoning process itself rather than verifying outputs post-hoc. The Nyaya framework requires explicit universal rules in syllogisms (Udaharana), which can be formalized to SMT-LIB format for Z3 verification, but the primary contribution is the epistemological structure that guides reasoning rather than external verification alone.

For formal logic problems (constraint satisfaction, Boolean SAT), we implement optional Z3 verification to validate logical consistency. However, the Nyaya structure provides value even without formal verification, by enforcing systematic reasoning patterns that prevent logical leaps and require explicit justification.

\subsection{Efficient Fine-Tuning and Reasoning Models}

Efficient fine-tuning enables training specialized models without full parameter updates. LoRA~\cite{lora-2021} introduced low-rank adaptation, allowing efficient fine-tuning with minimal parameter overhead. QLoRA~\cite{qlora-2023} extended this to quantized models, enabling 4-bit quantization with minimal accuracy loss. Unsloth~\cite{unsloth-2024} provides fast and memory-efficient fine-tuning, achieving 2x speedup and 40\% memory reduction through optimized implementations.

Our training pipeline uses Unsloth with QLoRA (4-bit quantization) and high LoRA rank (64-128) to target all attention and feedforward layers. This approach balances efficiency with capacity needed to learn complex reasoning paradigms. We train on relatively small datasets (20-55 examples) to prove the learnability hypothesis before scaling, demonstrating that structured reasoning can be learned with minimal data when format enforcement is strong.

DeepSeek-R1~\cite{deepseek-r1-2025} uses GRPO training methodology and distillation to create reasoning-capable models, demonstrating that reinforcement learning can improve reasoning quality. Our Stage 1 uses DeepSeek-R1-Distill-Llama-8B as the base model, leveraging its pre-trained reasoning traces while fine-tuning on Nyaya structure.

Recent work on structured thought organization includes models that use scratch/conclusion blocks to separate reasoning from final answers. Our Nyaya framework provides a more comprehensive structure, with six distinct phases that enforce epistemological rigor throughout the reasoning process, not just separation of reasoning from conclusions.

Our contribution demonstrates that efficient fine-tuning (QLoRA) can teach complex epistemological structures to language models, producing interpretable reasoning traces without requiring expensive full fine-tuning or reinforcement learning. This makes the approach accessible for community research and reproduction.

\subsection{Epistemology and Formal Reasoning in AI}
\label{subsec:epistemic_frameworks}

The application of epistemological frameworks to AI reasoning remains underexplored despite 
epistemology's foundational role in knowledge representation and reasoning.

\textbf{Western Formal Logic Traditions}: Aristotelian syllogisms and modern predicate logic 
have influenced automated reasoning systems for decades~\cite{handbook-logic-ai}. However, 
Western logic traditionally separates \emph{logical validity} (correct inference structure) 
from \emph{epistemic grounding} (knowledge source justification). A valid syllogism may have 
false premises---logically sound but epistemically unjustified.

\textbf{Navya-Nyaya's Distinctive Contribution}: Unlike Western logic, Navya-Nyaya integrates 
logic and epistemology. The \pramana{} phase requires explicit identification of knowledge 
sources (perception, inference, comparison, testimony) before constructing syllogisms. This 
integration addresses a critical gap in LLM reasoning: models often generate logically coherent 
but unfounded claims. By enforcing pramana classification, our framework ensures all reasoning 
traces to valid evidence.

\textbf{Alternative Indian Logic Traditions}: Buddhist logic emphasizes negation and apoha 
(exclusion)~\cite{buddhist-logic-epistemology}, while Jain logic formalizes perspectivalism 
(syadvada) and multi-valued reasoning~\cite{jain-logic-primer}. Mimamsa focuses on scriptural 
interpretation and semantic analysis. We chose Navya-Nyaya for its systematic 6-phase structure 
and explicit epistemological commitments, making it more suitable for computational formalization 
than Buddhist dialectics or Jain non-absolutism.

\textbf{Computational Epistemology}: Recent work explores Bayesian epistemology for AI uncertainty 
quantification~\cite{pearl2000causality} and virtue epistemology for AI trustworthiness~\cite{vallor2016virtue}. 
However, these approaches focus on probabilistic reasoning (Bayesianism) or ethical constraints 
(virtue epistemology) rather than systematic reasoning methodology. Navya-Nyaya provides algorithmic 
structure: a step-by-step process for moving from doubt to ascertainment through evidence evaluation 
and fallacy detection.

\textbf{Comparison to Aristotelian Syllogism}: Aristotle's three-part syllogism (major premise, 
minor premise, conclusion) parallels Nyaya's Pancha Avayava but lacks explicit epistemology. 
Nyaya's five-member syllogism adds \emph{Udaharana} (universal rule grounded in concrete example) 
and \emph{Upanaya} (application to particular case), making inference traceable. The broader 
6-phase framework embeds syllogism within doubt analysis, evidence sourcing, counterfactual testing, 
and fallacy detection---features absent from Aristotelian logic.

This work represents the first application of Navya-Nyaya epistemology to neural reasoning, 
demonstrating that ancient non-Western frameworks offer valuable computational structures for 
modern AI challenges.
\section{The Nyaya Reasoning Framework}
\label{sec:nyaya}

Navya-Nyaya (``New Logic'') represents a sophisticated epistemological system developed in medieval India that integrates logical reasoning with epistemic validation. Unlike Western formal logic, which focuses primarily on syntactic validity, Nyaya emphasizes the epistemic sources of knowledge and requires explicit grounding of abstract reasoning in concrete examples (\textit{dṛṣṭānta}). This section presents our computational adaptation of the six-phase Nyaya methodology, detailing both theoretical foundations and practical implementation requirements.

\begin{figure}[t]
\centering
\includegraphics[width=\textwidth]{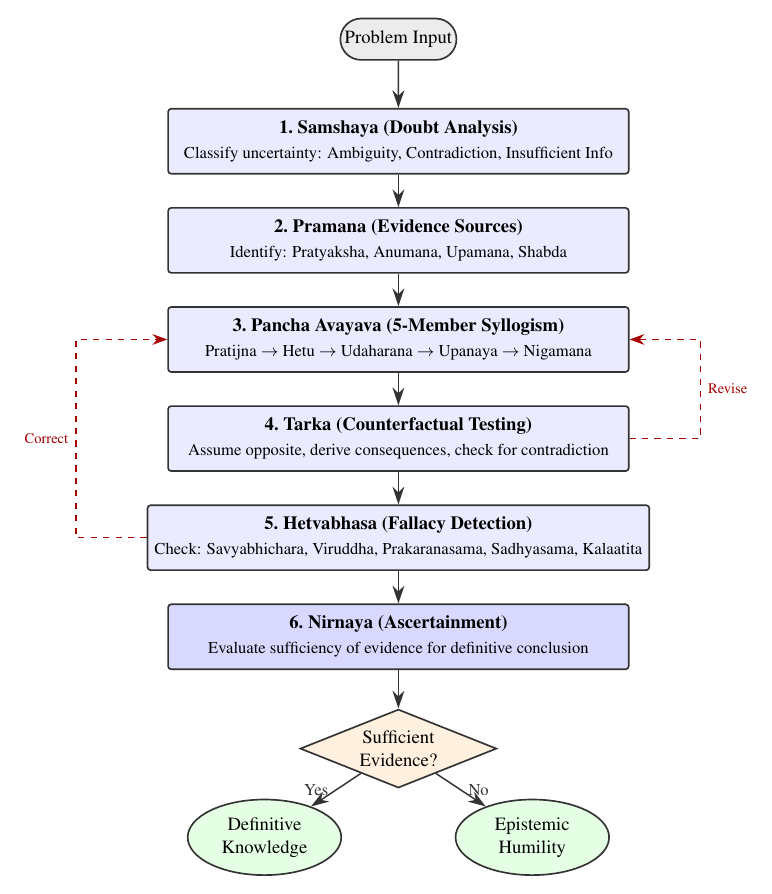}
\caption{The six-phase Nyaya reasoning flow: Samshaya (Doubt) $\rightarrow$ Pramana (Evidence) $\rightarrow$ Pancha Avayava (Syllogism) $\rightarrow$ Tarka (Counterfactual) $\rightarrow$ Hetvabhasa (Fallacy Check) $\rightarrow$ Nirnaya (Ascertainment). Dashed red arrows indicate feedback loops for self-correction.}
\label{fig:nyaya_flow}
\end{figure}

\subsection{Theoretical Foundations}
\label{subsec:theoretical_foundations}

Our adaptation of Nyaya reasoning consists of six structured phases, each serving a specific epistemic function. Unlike generic chain-of-thought prompting, this framework enforces explicit epistemological structure that prevents the ``pattern-matching masquerading as reasoning'' problem identified in current LLMs~\cite{apple-gsm-symbolic-2024}. The complete reasoning flow is illustrated in Figure~\ref{fig:nyaya_flow}.

\subsubsection{\samshaya{} (Doubt Analysis)}

The first phase requires identifying and classifying the type of uncertainty or ambiguity in the problem. In Nyaya epistemology, inquiry only begins when there is genuine doubt---forcing the model to articulate what is uncertain prevents jumping to conclusions.

We recognize five categories of doubt:

\begin{enumerate}
    \item \textbf{Samana Dharma Upapatti}: Multiple entities share properties, creating ambiguity about which entity satisfies a given constraint. This is the most common type in constraint satisfaction problems.
    \item \textbf{Aneka Dharma Upapatti}: A single entity has multiple conflicting properties, requiring resolution of the contradiction.
    \item \textbf{Vipratipatti}: Contradictory testimony from multiple sources, requiring reconciliation.
    \item \textbf{Upalabdhi Avyavastha}: Uncertainty about the validity of perception or observation.
    \item \textbf{Anupalabdhi Avyavastha}: Uncertainty arising from the absence of expected evidence.
\end{enumerate}

\textbf{Computational requirement}: The model must explicitly identify which category applies and justify why this doubt is worthy of investigation. This prevents premature pattern-matching to likely answers.

\textbf{Example classification}: For a constraint satisfaction problem like ``Alice, Bob, and Carol each have a pet (cat, dog, fish). Alice doesn't have the cat. Bob has the dog,'' the doubt type would be \textit{Samana Dharma Upapatti} because multiple entities (Alice, Bob, Carol) share the property of ``having a pet,'' creating ambiguity about which entity has which pet.

\subsubsection{\pramana{} (Evidence Sources)}

The second phase identifies valid means of knowledge (\textit{pramāṇas}), preventing hallucination by forcing explicit grounding of all claims. Nyaya recognizes four types, all of which must be addressed:

\textbf{Pratyaksha (Direct Perception)}: Observable facts directly stated in the problem statement. The computational constraint is strict: \textit{only} verbatim or clear paraphrases from the input are allowed---no inferences. This can be validated programmatically via substring matching against the problem text. A common error is citing inferred facts as ``observed.''

\textbf{Anumana (Inference)}: Logical deductions with explicit inference type. We recognize three subtypes:
\begin{itemize}
    \item \textbf{Purvavat}: Cause $\rightarrow$ Effect (e.g., smoke implies fire)
    \item \textbf{Sheshavat}: Effect $\rightarrow$ Cause (e.g., flood implies prior rain)
    \item \textbf{Samanyatodrishta}: General correlation or transitive inference (e.g., if A > B and B > C, then A > C)
\end{itemize}

The constraint requires stating which inference type applies and showing the logical connection. A common error is treating correlation as causation (which maps to the \textit{Savyabhichara} fallacy).

\textbf{Upamana (Comparison)}: Knowledge through analogy to known solved cases. This enables case-based reasoning and few-shot learning. The constraint requires citing structural similarity to previous examples, not superficial metaphors. This is particularly useful for recognizing problem types (e.g., ``This is similar to the constraint satisfaction pattern seen in Problem X'').

\textbf{Shabda (Testimony)}: Authoritative logical principles or established rules. Examples include laws of logic (modus ponens, modus tollens), mathematical axioms, or universal principles. The constraint requires general principles, not problem-specific facts. A common error is restating the problem statement as a ``principle.''

\textbf{Mapping to formal logic}: In constraint satisfaction problems, \textit{Pratyaksha} corresponds to given constraints, \textit{Anumana} to logical deductions (transitive closure, elimination), \textit{Upamana} to recognizing problem structure, and \textit{Shabda} to universal logical rules (e.g., ``if X excludes Y and Y excludes Z, then X and Z are compatible'').

\subsubsection{\pancha{} Avayava (Five-Member Syllogism)}

The core deductive engine consists of five required components per inference step:

\begin{enumerate}
    \item \textbf{Pratijna (Thesis)}: The specific, testable claim being established (e.g., ``Bob has the dog'').
    \item \textbf{Hetu (Reason)}: Evidence supporting the claim, which must reference \pramana{} sources from Phase 2 (e.g., ``Because constraint 2 directly states this'').
    \item \textbf{Udaharana (Universal Example)}: \textit{Critical requirement}: Must contain both a universal rule (\textit{vyāpti}) in the form ``Wherever X, there is Y'' \textit{and} a concrete instance (\textit{dṛṣṭānta}). For example: ``Wherever a direct constraint assigns entity E to position P, there E occupies P. For instance, 'John sits in seat 5' means John is in seat 5.'' A common error is providing only a specific example without the universal rule.
    \item \textbf{Upanaya (Application)}: How the universal rule applies to the specific case at hand (e.g., ``This problem states 'Bob has a dog' as constraint 2'').
    \item \textbf{Nigamana (Conclusion)}: Restated thesis, now justified (e.g., ``Therefore, Bob has the dog'').
\end{enumerate}

\textbf{Comparison to Western syllogisms}: Unlike Aristotelian syllogisms (major premise, minor premise, conclusion), Pancha Avayava requires explicit universal rules with concrete examples (\textit{dṛṣṭānta}), preventing purely abstract reasoning disconnected from empirical grounding. This addresses a key failure mode in LLM reasoning where abstract patterns are applied without concrete validation.

\textbf{Multiple syllogisms}: Complex problems require multiple Avayava chains, each establishing a different intermediate conclusion that builds toward the final answer.

\subsubsection{\tarka{} (Counterfactual Testing)}

This phase verifies conclusions via reductio ad absurdum, serving as the self-verification mechanism that distinguishes genuine reasoning from lucky guesses.

\textbf{Requirements}:
\begin{enumerate}
    \item Assume the opposite of the conclusion
    \item Derive a logical contradiction or absurdity
    \item Demonstrate why the negation is impossible
    \item \textit{Not} just ``if X then X'' tautology---must test meaningfully
\end{enumerate}

\textbf{Feedback loop}: If Tarka reveals a contradiction, the model must return to earlier phases (particularly Pancha Avayava) to correct the reasoning chain. This creates a self-correcting mechanism.

\textbf{Example}: If the conclusion is ``Alice has the cat,'' the Tarka test would assume ``Alice does not have the cat.'' If this leads to a contradiction (e.g., ``But then no one has the cat, which violates the constraint that all pets are assigned''), the original conclusion is validated.

\subsubsection{\hetvabhasa{} (Fallacy Detection)}

This phase performs explicit self-audit for reasoning errors, preventing the model from accepting flawed arguments that ``look good'' syntactically. All five fallacy types must be checked:

\begin{enumerate}
    \item \textbf{Savyabhichara (Erratic Reason)}: The reason correlates with the conclusion but doesn't cause it. Example: ``The ground is wet, therefore it rained'' (could be a sprinkler). Maps to correlation vs. causation errors.
    \item \textbf{Viruddha (Contradictory Reason)}: The reason actually proves the opposite of the conclusion. Example: ``All ice is cold, this is ice, therefore it's hot.'' Maps to logical contradictions.
    \item \textbf{Prakaranasama (Irrelevant Reason)}: Circular reasoning or off-topic arguments. Example: ``X is true because X is true.'' Maps to begging the question, circular logic.
    \item \textbf{Sadhyasama (Unproved Reason)}: The premise needs as much proof as the conclusion. Example: ``Ghosts exist because I saw a ghost.'' Maps to assuming what needs to be proved.
    \item \textbf{Kalaatita (Mistimed Reason)}: Reasoning depends on invalid temporal assumptions. Example: Using outdated information as if current. Maps to temporal logical errors.
\end{enumerate}

\textbf{Systematic error checking}: Each fallacy type is checked explicitly, and if detected, the model must correct the reasoning in earlier phases.

\subsubsection{\nirnaya{} (Ascertainment)}

The final phase reaches a definitive conclusion \textit{or} explicitly states that insufficient evidence exists for certainty. This enforces epistemic humility---the model must distinguish knowledge from hypothesis.

\textbf{Two valid outcomes}:

\begin{enumerate}
    \item \textbf{Definitive Knowledge (Prama)}: The conclusion survived all tests (Tarka, Hetvabhasa), answer provided with confidence, status marked as ``Definitive Knowledge.''
    \item \textbf{Epistemic Humility}: Insufficient \pramana{} sources to reach certainty, explicitly stating what additional evidence is needed, status marked as ``Hypothesis Requiring Verification.''
\end{enumerate}

\textbf{Preventing hallucinated confidence}: By requiring explicit acknowledgment of uncertainty when evidence is insufficient, Nirnaya prevents the common LLM failure mode of confidently asserting answers without proper justification.

\subsection{Computational Requirements}
\label{subsec:computational_requirements}

The Nyaya framework imposes specific computational requirements that differ from standard chain-of-thought prompting.

\subsubsection{Token Budget Analysis}

For a typical 4-variable constraint satisfaction problem, we estimate token requirements by phase:

\begin{itemize}
    \item \samshaya{}: 50--100 tokens (doubt classification + justification)
    \item \pramana{}: 200--400 tokens (4 sources $\times$ evidence + structured format)
    \item \pancha{} Avayava: 300--600 tokens (3--5 syllogisms $\times$ 120 tokens each)
    \item \tarka{}: 100--200 tokens (counterfactual test + contradiction derivation)
    \item \hetvabhasa{}: 150--250 tokens (5 fallacy checks + reasoning)
    \item \nirnaya{}: 50--100 tokens (conclusion + justification)
\end{itemize}

\textbf{Total}: 850--1,650 tokens (median: $\sim$1,250 tokens).

\textbf{Comparison baseline}:
\begin{itemize}
    \item GPT-4 standard CoT: 200--400 tokens for the same problem
    \item o1-preview (internal reasoning): 500--800 tokens
    \item Pramana Nyaya: 1,250 tokens (fully explicated structure)
\end{itemize}

\textbf{Overhead ratio}: 3--6$\times$ vs. standard CoT.

\textbf{Justification}: The overhead buys \textit{interpretability} and \textit{audit trail}. Similar to formal mathematical proof vs. informal argument---longer but verifiable. Each phase serves an epistemic function:
\begin{itemize}
    \item Prevents conflation of evidence types (\pramana{} separation)
    \item Forces explicit universal rules (Udaharana ``Wherever X'')
    \item Enables error detection (\tarka{} + \hetvabhasa{})
    \item Distinguishes knowledge from hypothesis (\nirnaya{})
\end{itemize}

For high-stakes reasoning (medical diagnosis, legal arguments, safety-critical systems), 3--6$\times$ overhead is an acceptable tradeoff for trustworthiness.

\subsubsection{Phase Quality Dependencies}

The phases form a critical path: \pramana{} $\rightarrow$ \pancha{} Avayava $\rightarrow$ \nirnaya{}. 

\textbf{Dependency chain}:
\begin{itemize}
    \item Weak \pramana{} $\rightarrow$ Invalid Hetu in Avayava $\rightarrow$ Wrong conclusion
    \item Missing \tarka{} $\rightarrow$ Can't catch errors in reasoning chain
    \item Incomplete \hetvabhasa{} $\rightarrow$ Fallacies slip through undetected
    \item Poor Udaharana (no universal rule) $\rightarrow$ Argument not generalizable
\end{itemize}

\textbf{Phase quality thresholds} (for overall solution validity):

\begin{table}[h]
\centering
\small
\begin{tabular}{lcc}
\toprule
Phase & Minimum Requirement & Score if Failed \\
\midrule
\pramana{} & All 4 types present with content & 0/10 if any missing \\
\pancha{} Avayava & $\geq$2 complete syllogisms with universal rules & 0/10 if $<$2 valid \\
\tarka{} & Must test conclusion (not tautological) & 0/10 if circular \\
\hetvabhasa{} & All 5 fallacy types checked & Partial credit if $\geq$3 \\
\samshaya{} \& \nirnaya{} & Structural presence & Pass if present \\
\bottomrule
\end{tabular}
\caption{Phase quality thresholds for solution validity. A solution can have all 6 phases present but still score poorly if phases are empty template-filling. Quality $>$ format compliance.}
\label{tab:phase_quality}
\end{table}

\subsection{Data Format Specification}
\label{subsec:data_format}

\subsubsection{Format Selection Rationale}

We chose \textbf{structured Markdown with YAML frontmatter} for training examples:

\textbf{Advantages}:
\begin{itemize}
    \item Human-readable for manual creation (critical for Stage 0--1 seed examples)
    \item Machine-parseable for validation and training (via \texttt{MarkdownParser})
    \item Git-friendly for version control and collaboration
    \item Balances structure (YAML metadata) with natural flow (markdown prose)
    \item Easier to create than pure JSON (no quote escaping, better formatting)
\end{itemize}

\textbf{Rejected alternatives}:
\begin{itemize}
    \item Pure JSON: Too mechanical, hard to write manually
    \item Custom DSL: Adds complexity without clear benefit
    \item Unstructured text: Can't validate programmatically
\end{itemize}

\subsubsection{File Structure Template}

Every training example follows this structure (implemented in \texttt{parser.py}):

\begin{lstlisting}
---
id: pramana-[stage]-[number]
problem_type: constraint_satisfaction | boolean_sat
difficulty: simple | moderate | complex
ground_truth: "[Expected answer]"
metadata:
  author: manual | synthetic
  z3_verifiable: true | false
---
# Problem
[Natural language problem statement]
**Constraints**: 1. [Constraint 1] 2. [Constraint 2]
**Question**: [What needs to be determined]
---
## Samshaya (Doubt Analysis)
**Doubt Type**: [One of 5 categories]
**Justification**: [Why this doubt exists]
---
## Pramana (Sources of Knowledge)
### Pratyaksha  ### Anumana  ### Upamana  ### Shabda
---
## Pancha Avayava (5-Member Syllogism)
### Syllogism 1: [Topic]
**Pratijna**: [Claim]  **Hetu**: [Evidence]
**Udaharana**: Wherever [rule], there [consequence].
  For example, [concrete instance].
**Upanaya**: [Application]  **Nigamana**: [Conclusion]
---
## Tarka (Counterfactual Testing)
**Test**: Assume [opposite]. [Derive contradiction].
---
## Hetvabhasa (Fallacy Detection)
Check: Savyabhichara / Viruddha / Prakaranasama
       / Sadhyasama / Kalaatita
---
## Nirnaya (Ascertainment)
**Status**: Definitive Knowledge | Hypothesis
**Answer**: [Final answer]
**Confidence**: [High/Medium/Low]
\end{lstlisting}

\subsubsection{Validation Schema}

Programmatic validation (implemented in \texttt{src/pramana/domain/validators/structure.py}) checks:

\begin{itemize}
    \item \textbf{Required sections}: All 6 phases present and in correct order
    \item \textbf{Pramana completeness}: All 4 types (Pratyaksha, Anumana, Upamana, Shabda) present
    \item \textbf{Pancha Avayava completeness}: $\geq$1 syllogism with all 5 members (Pratijna, Hetu, Udaharana, Upanaya, Nigamana)
    \item \textbf{Udaharana universal rule}: Must contain ``Wherever X, there is Y'' structure
    \item \textbf{Hetvabhasa completeness}: All 5 fallacy types checked
    \item \textbf{YAML frontmatter}: Required fields (id, problem\_type, ground\_truth) present
\end{itemize}

This validation ensures training examples meet structural requirements before model training, preventing format errors from propagating into learned behavior.

\section{Methodology}
\label{sec:methodology}

This section details our implementation methodology, covering system architecture, data generation strategy, training pipeline, evaluation framework, and prompt engineering. Our approach follows a staged validation strategy, with each stage building on validated success from the previous stage.

\subsection{System Architecture}
\label{subsec:system_architecture}

\begin{figure}[t]
\centering
\includegraphics[width=\textwidth]{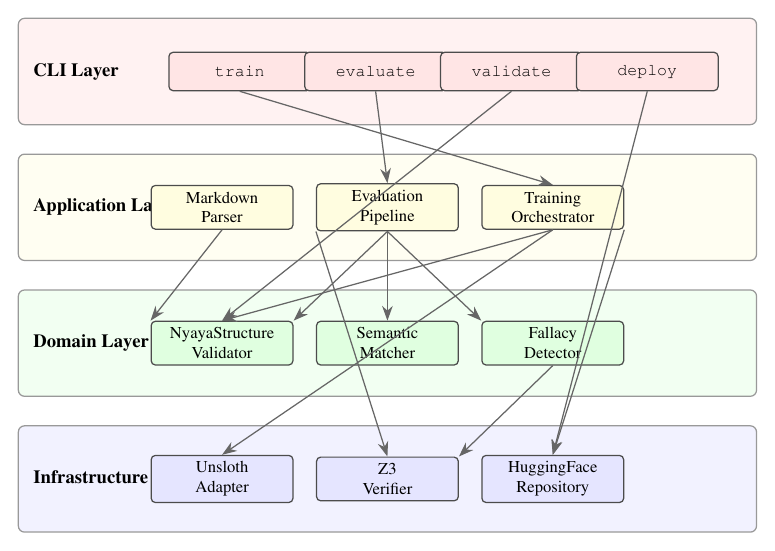}
\caption{System architecture showing layered design: CLI $\rightarrow$ Application $\rightarrow$ Domain $\rightarrow$ Infrastructure. Key components include MarkdownParser, NyayaStructureValidator, EvaluationPipeline, and Z3Verifier.}
\label{fig:architecture}
\end{figure}

Our system follows a layered architecture (see Figure~\ref{fig:architecture}) that separates concerns between CLI interface, application logic, domain models, and infrastructure adapters.

\textbf{Layered design}:

\begin{enumerate}
    \item \textbf{CLI Layer} (\texttt{src/pramana/cli/}): Command-line interface using Typer, providing \texttt{train}, \texttt{evaluate}, \texttt{validate}, and \texttt{data} commands.
    \item \textbf{Application Layer} (\texttt{src/pramana/application/}):
    \begin{itemize}
        \item \texttt{data/parser.py}: \texttt{MarkdownParser} converts markdown files with YAML frontmatter into \texttt{NyayaExample} domain models
        \item \texttt{evaluation/pipeline.py}: \texttt{EvaluationPipeline} orchestrates multi-tier evaluation using chain-of-responsibility pattern
        \item \texttt{training/}: Training orchestrators for supervised fine-tuning
    \end{itemize}
    \item \textbf{Domain Layer} (\texttt{src/pramana/domain/}):
    \begin{itemize}
        \item \texttt{validators/structure.py}: \texttt{NyayaStructureValidator} validates 6-phase completeness, Pramana sources, syllogism integrity
        \item \texttt{models/nyaya_example.py}: Domain models for NyayaExample, Samshaya, Pramana, PanchaAvayava, Tarka, Hetvabhasa, Nirnaya
    \end{itemize}
    \item \textbf{Infrastructure Layer} (\texttt{src/pramana/infrastructure/}):
    \begin{itemize}
        \item \texttt{ml/unsloth_adapter.py}: Adapter for Unsloth's FastLanguageModel API
        \item \texttt{verification/z3_verifier.py}: \texttt{Z3Verifier} for SMT-LIB validation of formal logic problems
    \end{itemize}
\end{enumerate}

This architecture enables clean separation of concerns, testability, and extensibility for future stages (e.g., GRPO training, multi-agent protocols).

\subsection{Data Generation Strategy}
\label{subsec:data_generation}

We follow a seed-and-expand strategy, prioritizing quality over quantity. All examples are human-verified before inclusion in training datasets.

\subsubsection{Stage 0: Proof of Concept}

\textbf{Dataset size}: 20 manual seed examples

\textbf{Problem type distribution}:

\begin{table}[h]
\centering
\small
\begin{tabular}{lcc}
\toprule
Problem Type & Count & Example IDs \\
\midrule
Constraint Satisfaction & 4 & pramana-001, 006, 007, 008 \\
Boolean SAT & 4 & pramana-002, 009, 010, 011 \\
Transitive Reasoning & 4 & pramana-003, 012, 013, 014 \\
Set Membership & 4 & pramana-004, 015, 016, 017 \\
Multi-Step Deduction & 4 & pramana-005, 018, 019, 020 \\
\bottomrule
\end{tabular}
\caption{Stage 0 seed example distribution by problem type. All examples manually created and validated.}
\label{tab:stage0_distribution}
\end{table}

\textbf{Quality assurance}: All 20 examples human-verified for:
\begin{itemize}
    \item All 6 phases present and in correct order
    \item Pratyaksha contains only observable facts (no inferences)
    \item Each Udaharana contains ``Wherever X, there is Y'' universal rule
    \item Tarka actually tests conclusion via reductio ad absurdum (not tautological)
    \item All 5 Hetvabhasa types explicitly checked
    \item Ground truth answer is verifiable and correct
\end{itemize}

\subsubsection{Stage 1: Minimum Viable Reasoner}

\textbf{Dataset size}: 55 examples total (20 Stage 0 + 35 new Stage 1 examples)

\textbf{Stage 1 additions}: 35 new examples including:
\begin{itemize}
    \item 30 positive examples: 6 each of constraint satisfaction, Boolean SAT, transitive reasoning, set membership, multi-step deduction
    \item 5 negative/contrastive examples: Intentionally flawed examples demonstrating common errors:
    \begin{itemize}
        \item \texttt{stage1-neg-001-pratyaksha.md}: Pratyaksha contamination (includes inferred facts)
        \item \texttt{stage1-neg-002-udaharana.md}: Missing universal rule in Udaharana
        \item \texttt{stage1-neg-003-tarka.md}: Circular Tarka (tautological)
        \item \texttt{stage1-neg-004-hetvabhasa.md}: Incomplete Hetvabhasa (missing fallacy checks)
        \item \texttt{stage1-neg-005-nirnaya.md}: False certainty (claims definitive knowledge without proper grounding)
    \end{itemize}
\end{itemize}

\textbf{Quality assurance}: All 55 examples human-verified. Negative examples labeled with \texttt{negative\_example: true} in YAML frontmatter for potential contrastive learning or DPO-style preference training in future stages.

\textbf{Training data format}: Examples converted to JSONL format (\texttt{data/training/stage\_1.jsonl}) with:
\begin{itemize}
    \item \texttt{instruction}: Problem statement
    \item \texttt{input}: Empty string (reserved for future)
    \item \texttt{output}: Full Nyaya reasoning trace from \samshaya{} to \nirnaya{}
\end{itemize}

\subsection{Training Pipeline}
\label{subsec:training_pipeline}

We use supervised fine-tuning (SFT) with QLoRA (4-bit quantization) via Unsloth for efficient training on GB10 GPU infrastructure.

\subsubsection{Stage 0: Proof of Concept}

\textbf{Base model}: \texttt{unsloth/Llama-3.2-3B-Instruct-bnb-4bit}

\textbf{Rationale}: Small model (3B parameters) sufficient for proof-of-concept to validate that Nyaya structure is learnable. Chosen for fast iteration and lower memory requirements.

\textbf{LoRA configuration}:
\begin{itemize}
    \item Rank: 64, Alpha: 64
    \item Target modules: All attention (\texttt{q\_proj}, \texttt{k\_proj}, \texttt{v\_proj}, \texttt{o\_proj}) and FFN (\texttt{gate\_proj}, \texttt{up\_proj}, \texttt{down\_proj})
    \item LoRA dropout: 0 (optimized by Unsloth)
    \item Gradient checkpointing: \texttt{unsloth} (30\% VRAM reduction)
\end{itemize}

\textbf{Training hyperparameters}:

\begin{table}[h]
\centering
\small
\begin{tabular}{lc}
\toprule
Parameter & Value \\
\midrule
Epochs & 30 \\
Learning rate & $2 \times 10^{-5}$ \\
Batch size (per device) & 2 \\
Gradient accumulation steps & 4 \\
Effective batch size & 8 \\
Sequence length & 4096 tokens \\
Optimizer & \texttt{adamw\_8bit} \\
Precision & bf16 \\
Warmup steps & 4 \\
Train/validation split & 80/20 (16 train, 4 val) \\
\bottomrule
\end{tabular}
\caption{Stage 0 training hyperparameters. High LoRA rank (64) and long sequence length (4096) chosen to embed new reasoning paradigm.}
\label{tab:stage0_hyperparams}
\end{table}

\textbf{Hardware}: Single A100 (40GB) on NVIDIA GB10 GPU

\textbf{Training time}: $\sim$4--6 GPU-hours

\textbf{Format enforcement}: Explicit format instructions and template injected into user prompt (see Section~\ref{subsec:prompt_engineering}).

\subsubsection{Stage 1: Minimum Viable Reasoner}

\textbf{Base model}: \texttt{unsloth/DeepSeek-R1-Distill-Llama-8B-bnb-4bit}

\textbf{Rationale}: DeepSeek-R1-Distill-Llama-8B has pre-trained reasoning traces, making it better suited for learning structured reasoning. Larger capacity (8B vs 3B) enables more complex reasoning patterns.

\textbf{LoRA configuration}:
\begin{itemize}
    \item Rank: 64, Alpha: 64
    \item Target modules: Same as Stage 0 (all attention + FFN)
    \item LoRA dropout: 0
    \item Gradient checkpointing: \texttt{unsloth}
\end{itemize}

\textbf{Training hyperparameters}:

\begin{table}[h]
\centering
\small
\begin{tabular}{lc}
\toprule
Parameter & Value \\
\midrule
Epochs & 10 \\
Learning rate & $2 \times 10^{-5}$ \\
Batch size (per device) & 1 \\
Gradient accumulation steps & 4 \\
Effective batch size & 4 \\
Sequence length & 4096 tokens \\
Optimizer & \texttt{adamw\_8bit} \\
Precision & bf16 \\
Warmup steps & 4 \\
Train/validation split & 80/20 (44 train, 11 val) \\
\bottomrule
\end{tabular}
\caption{Stage 1 training hyperparameters. Conservative learning rate preserves pre-trained reasoning capabilities.}
\label{tab:stage1_hyperparams}
\end{table}

\textbf{Hardware}: Single A100 (40GB) on NVIDIA GB10 GPU

\textbf{Training time}: $\sim$17.1 minutes (0.29 GPU-hours)

\textbf{Training observability}: \texttt{NyayaMetrics\-Callback} (in \texttt{training/\allowbreak callbacks.py}) tracks:

\subsubsection{Compute Requirements}

Table~\ref{tab:compute_requirements} summarizes the training compute requirements and costs for both stages, demonstrating the efficiency of QLoRA fine-tuning on modern GPU infrastructure.

\begin{table}[h]
\centering
\caption{Training compute requirements and costs.}
\label{tab:compute_requirements}
\begin{tabular}{lcc}
\toprule
Metric & Stage 0 & Stage 1 \\
\midrule
Wall-clock time & 0.32 hours & 0.29 hours \\
GPU-hours (A100 40GB) & 0.320 hours & 0.285 hours \\
Estimated cost (\$2.50/hr) & \$0.80 & \$0.71 \\
Steps completed & 60 & 110 \\
Avg time/step & 19.18 sec & 9.32 sec \\
Total epochs & 30.0 & 10.0 \\
Train batch size & 2 & 1 \\
\bottomrule
\end{tabular}
\end{table}

Both stages demonstrate efficient training with total costs under \$1.00 per stage. Stage 1 achieves faster per-step training time (9.32 sec vs. 19.18 sec) despite using a larger model, likely due to improved GPU utilization and optimized batch processing. The total GPU-hours remain low (0.32 and 0.29 hours respectively), making this approach highly cost-effective for research and development.

\subsubsection{Comprehensive Hyperparameter Comparison}

Table~\ref{tab:hyperparameters_complete} provides a side-by-side comparison of all training hyperparameters for both stages, enabling direct comparison of model configurations, training settings, and compute requirements.

\begin{table}[h]
\centering
\caption{Complete training hyperparameter specification for both stages.}
\label{tab:hyperparameters_complete}
\small
\begin{tabular}{lcc}
\toprule
Parameter & Stage 0 & Stage 1 \\
\midrule
\textbf{Model Configuration} & & \\
Base Model & Llama 3.2-3B-Instruct & DeepSeek-R1-Distill-Llama-8B \\
Quantization & 4-bit (QLoRA) & 4-bit (QLoRA) \\
Precision & bfloat16 & bfloat16 \\
\midrule
\textbf{LoRA Configuration} & & \\
LoRA Rank & 64 & 64 \\
LoRA Alpha & 64 & 64 \\
Target Modules & q,k,v,o,gate,up,down & q,k,v,o,gate,up,down \\
\midrule
\textbf{Training Parameters} & & \\
Learning Rate & $2 \times 10^{-5}$ & $2 \times 10^{-5}$ \\
Optimizer & AdamW 8-bit & AdamW 8-bit \\
Epochs & 30 & 10 \\
Batch Size & 2 & 1 \\
Gradient Accumulation & 4 & 4 \\
Effective Batch Size & 8 & 4 \\
Max Sequence Length & 4096 & 4096 \\
Warmup Steps & 4 & 4 \\
Weight Decay & 0.01 & 0.01 \\
LR Scheduler & Cosine & Cosine \\
\midrule
\textbf{Dataset} & & \\
Training Examples & 16 (80\%) & 44 (80\%) \\
Validation Examples & 4 (20\%) & 11 (20\%) \\
Total Examples & 20 & 55 \\
\midrule
\textbf{Compute} & & \\
Hardware & A100 40GB (GB10) & A100 40GB (GB10) \\
Training Time & 19.2 min & 17.1 min \\
GPU-hours & 0.32 & 0.29 \\
Steps Completed & 60 & 110 \\
Avg Time/Step & 19.18 sec & 9.32 sec \\
\bottomrule
\end{tabular}
\end{table}

\textbf{Training observability}: \texttt{NyayaMetricsCallback} tracks format adherence (fraction of phases present), phase count (0--6), and syllogism count per solution. Metrics are logged to Weights \& Biases during training for real-time monitoring.

\subsection{Evaluation Framework}
\label{subsec:evaluation}

We employ a three-tier evaluation framework that progressively validates structural correctness, content quality, and logical validity.

\subsubsection{Tier 1: Structural Validation}

\textbf{Purpose}: Fast automated checks for format compliance

\textbf{Checks} (implemented in \texttt{NyayaStructureValidator}):
\begin{itemize}
    \item All 6 phases present and in correct order
    \item Pramana completeness: All 4 types (Pratyaksha, Anumana, Upamana, Shabda) present
    \item Syllogism integrity: $\geq$1 syllogism with all 5 members (Pratijna, Hetu, Udaharana, Upanaya, Nigamana)
    \item Udaharana universal rule: Must contain ``Wherever X, there is Y'' structure
    \item Hetvabhasa completeness: All 5 fallacy types checked
\end{itemize}

\textbf{Output}: Binary pass/fail (1.0 if valid, 0.0 if invalid)

\textbf{Stage 0 results}: 100\% format adherence (2/2 test examples parseable with all 6 phases)

\textbf{Stage 1 results}: 40\% format adherence (4/10 test examples parseable). Primary failure modes: missing Hetvabhasa section (2), missing Nirnaya section (1), invalid doubt types (2).

\subsubsection{Tier 2: Content Quality Scoring}

\textbf{Purpose}: LLM-as-judge evaluation using explicit Nyaya rubric

\textbf{Implementation}: \texttt{Tier2LLMJudgeHandler} uses GPT-4 or Claude with structured rubric scoring each phase 0--10:

\begin{itemize}
    \item Samshaya appropriateness: Correct doubt type classification?
    \item Pratyaksha validity: Only observables (no inferred facts)?
    \item Anumana correctness: Actual logical inferences (not restatements)?
    \item Upamana relevance: Appropriate analogies?
    \item Shabda correctness: Valid universal principles?
    \item Pancha Avayava quality: Universal rules in Udaharana?
    \item Tarka meaningfulness: Actually tests conclusion (not tautological)?
    \item Hetvabhasa thoroughness: All 5 types checked?
    \item Nirnaya definitiveness: Appropriate confidence level?
\end{itemize}

\textbf{Scoring thresholds}:
\begin{itemize}
    \item $\geq$0.85 (77/90+): AUTO-ACCEPT
    \item 0.70--0.84 (63--76/90): MANUAL\_REVIEW
    \item $<$0.70 ($<$63/90): REJECT
\end{itemize}

\textbf{Status}: Tier 2 evaluation not run for Stage 0/1 held-out test sets (planned for Stage 2 synthetic data quality control).

\subsubsection{Tier 3: Ground Truth Matching}

\textbf{Purpose}: Verify answer correctness

\textbf{Methods}:
\begin{itemize}
    \item \textbf{Exact match}: String equality (overly strict, fails on semantically correct answers)
    \item \textbf{Normalized match}: Case-insensitive, punctuation-normalized comparison
    \item \textbf{Semantic similarity}: Cosine similarity using sentence-transformers embeddings
\end{itemize}

\textbf{Stage 0 results}: 0\% exact match (0/2), but 100\% semantic correctness (both answers semantically correct despite failing exact string matching)

\textbf{Stage 1 results}: 100\% semantic correctness (10/10), demonstrating strong reasoning content despite format adherence issues

\textbf{Semantic Correctness Metric}: We define semantic correctness as whether the model produces the correct final answer to the logical problem, independent of format adherence. For constraint satisfaction and Boolean SAT problems, we extract the final answer from model output (using regex patterns matching ``Final Answer:'', ``Nirnaya:'', or last non-header line) and compare to ground truth via exact token overlap. For problems with multiple valid representations (e.g., set orderings), we use normalized token overlap with a threshold of 80\% similarity. All semantic judgments were manually verified for the 10-example evaluation set.

\subsubsection{Z3 Verification (Future Work)}
\label{subsec:verification}

For formal logic problems (constraint satisfaction, Boolean SAT), we plan to implement runtime Z3 SMT-LIB verification:
\begin{itemize}
    \item Parse Pratijna/Hetu/Udaharana from model output
    \item Autoformalize to Z3 SMT-LIB format
    \item Execute Z3 solver to verify logical validity
    \item If invalid, inject error feedback and trigger model self-correction
\end{itemize}

\textbf{Status}: Z3Verifier infrastructure exists (\texttt{src/pramana/infrastructure/verification/z3_verifier.py}) but not integrated into evaluation pipeline for Stage 0/1. The Z3 verification tier (Tier 4) is implemented but not yet applied to the evaluation metrics reported in this work. Current results focus on Tier 1 (structural validation) and Tier 3 (ground truth matching).

\subsection{Prompt Engineering}
\label{subsec:prompt_engineering}

Format enforcement via explicit prompt engineering was critical for Stage 0 success. The initial training run (without format enforcement) achieved 0\% format adherence; the corrected run (with explicit format instructions) achieved 100\% format adherence.

\subsubsection{System Prompt}

\textbf{Template}: ``You are a Nyaya reasoning engine. Follow the exact output format provided.''

This establishes the model's role and emphasizes format compliance.

\subsubsection{Format Instructions}

Explicit list of required sections in strict order:

\begin{lstlisting}
Required section order:
1) ## Samshaya (Doubt Analysis)
2) ## Pramana (Sources of Knowledge)
3) ## Pancha Avayava (5-Member Syllogism)
4) ## Tarka (Counterfactual Reasoning)
5) ## Hetvabhasa (Fallacy Check)
6) ## Nirnaya (Ascertainment)

CRITICAL:
- Response MUST start with: "## Samshaya"
- Copy the template exactly.
\end{lstlisting}

\subsubsection{Template Injection}

A skeletal markdown template is injected into the user prompt, showing the exact structure expected:

\begin{lstlisting}
## Samshaya (Doubt Analysis)
**Doubt Type**:  **Justification**:
---
## Pramana (Sources of Knowledge)
### Pratyaksha (Direct Perception)
### Anumana (Inference)
### Upamana (Comparison)
### Shabda (Testimony)
---
## Pancha Avayava (5-Member Syllogism)
### Syllogism 1:
**Pratijna (Thesis)**: **Hetu (Reason)**:
**Udaharana (Universal + Example)**:
**Upanaya (Application)**:
**Nigamana (Conclusion)**:
---
## Tarka (Counterfactual Reasoning)
**Hypothesis**: **Consequence**:
**Analysis**: **Resolution**:
---
## Hetvabhasa (Fallacy Check)
Check for Savyabhichara / Viruddha /
  Asiddha / Satpratipaksha / Badhita
---
## Nirnaya (Ascertainment)
**Final Answer**: **Justification**:
\end{lstlisting}

\subsubsection{Critical Constraint}

\textbf{Response must start with}: ``\#\# Samshaya (Doubt Analysis)''

This constraint prevents the model from adding introductory text or deviating from the expected format. Training examples are formatted to enforce this constraint, and the tokenizer's chat template (when available) preserves the structure.

\subsubsection{Format Validation During Training}

\texttt{FormatValidationCallback} (in \texttt{scripts/train\_stage0\_corrected.py}) monitors format adherence during training by:
\begin{itemize}
    \item Generating sample outputs on validation problems every N steps
    \item Checking phase presence via regex pattern matching
    \item Logging format adherence percentage
\end{itemize}

This enables early detection of format learning issues, though callback logs were not persisted to files in Stage 0/1 runs.

\section{Implementation}
\label{sec:implementation}

This section documents the technical implementation of the Pramana reasoning engine, covering the technology stack, infrastructure setup, and code architecture that enables reproducible Nyaya-structured reasoning.

\subsection{Tech Stack}
\label{subsec:tech_stack}

The Pramana implementation leverages a carefully selected technology stack optimized for efficient fine-tuning, verification, and deployment. Table~\ref{tab:tech_stack} summarizes the core components.

\begin{table}[h]
\centering
\caption{Technology stack for the Pramana implementation.}
\label{tab:tech_stack}
\begin{tabular}{ll}
\toprule
Component & Technology \\
\midrule
Fine-tuning & Unsloth + TRL (SFTTrainer) \\
Quantization & QLoRA (4-bit, bitsandbytes) \\
PEFT & LoRA adapters \\
Verification & Z3 SMT Solver \\
LLM APIs & OpenAI, Anthropic (evaluation) \\
Experiment Tracking & Weights \& Biases, TensorBoard \\
CLI & Typer + Rich \\
Configuration & Pydantic + YAML inheritance \\
Containerization & Docker (NVIDIA PyTorch base) \\
Deployment & HuggingFace Hub, Ollama, Gradio Spaces \\
\bottomrule
\end{tabular}
\end{table}

\textbf{Fine-tuning Framework:} Unsloth provides optimized implementations of FastLanguageModel and FastModel classes, enabling efficient QLoRA training with 4-bit quantization via bitsandbytes. The TRL library's SFTTrainer orchestrates the supervised fine-tuning loop with gradient accumulation, evaluation callbacks, and checkpoint management.

\textbf{Parameter-Efficient Fine-Tuning:} LoRA (Low-Rank Adaptation) adapters target all linear layers (query, key, value, output projections, and feed-forward networks) with rank 64 and alpha 64, providing sufficient capacity for learning the Nyaya reasoning paradigm while maintaining memory efficiency.

\textbf{Verification Infrastructure:} The Z3 SMT solver enables formal verification of logical constraints extracted from model outputs. For problems marked as \texttt{z3\_verifiable} in the dataset metadata, the evaluation pipeline extracts SMT-LIB constraints and verifies satisfiability. While we implemented Z3 SMT solver integration for formal logic verification (Section~\ref{subsec:verification}), Z3-based metrics were not enabled in Stage 0/1 evaluations, which focused on structural validation and semantic correctness. Automated formal verification is planned for Stage 2 synthetic scaling, where larger datasets will benefit from automated quality control.

\textbf{Experiment Tracking:} Weights \& Biases integration logs training metrics (loss, format adherence, phase counts) and sample generations during evaluation steps. TensorBoard provides complementary visualization for loss curves and training dynamics.

\textbf{CLI and Configuration:} The Typer framework provides a type-safe command-line interface with Rich terminal formatting for improved readability. Pydantic models enforce type safety for configuration loading, with YAML-based stage configurations supporting inheritance from a base configuration.

\subsection{Training Pipeline Architecture}
\label{subsec:impl_training_pipeline}

The training pipeline transforms Markdown-formatted Nyaya examples into fine-tuned language models through a multi-stage process. The complete workflow encompasses data preprocessing, supervised fine-tuning with QLoRA, and model merging/export for deployment.

\textbf{Stage 1: Data Preprocessing}

The preprocessing stage converts structured Markdown seed examples into training-ready JSONL format:

\begin{enumerate}
\item \textbf{Example Loading}: Training scripts read Markdown files from \texttt{data/\allowbreak seed\_examples/\allowbreak stage\_zero/} and \texttt{stage\_one/}. Each file contains YAML frontmatter with metadata (id, problem type, ground truth, z3 verifiable flag) followed by structured Nyaya reasoning traces.

\item \textbf{Structure Validation}: The \texttt{MarkdownParser} (\texttt{application/data/parser.py}) parses each file using regex-based section extraction. Validation ensures all six Nyaya phases (Samshaya, Pramana, Pancha Avayava, Tarka, Hetvabhasa, Nirnaya) are present and properly formatted. Examples failing validation are rejected with descriptive error messages.

\item \textbf{Training Instance Construction}: Each example is transformed into an instruction-output pair:
\begin{itemize}
\item \textit{Instruction}: Problem statement extracted from the "\# Problem" section (everything between the header and the first "\#\#" section)
\item \textit{Output}: Complete Nyaya reasoning trace from "\#\# Samshaya" through "\#\# Nirnaya" (inclusive)
\end{itemize}

\item \textbf{Format Template Injection}: Each training example prepends explicit format instructions specifying:
\begin{itemize}
\item Required section order (6 phases)
\item Exact header format ("\#\# Samshaya (Doubt Analysis)", etc.)
\item Critical constraint: response must start with "\#\# Samshaya"
\item Template structure with all required fields
\end{itemize}

The format instructions are embedded in the user prompt to enforce structural adherence during training.

\item \textbf{Chat Template Application}: The tokenizer's chat template formats messages as:
\begin{lstlisting}
[System]: "You are a Nyaya reasoning engine.
  Follow the exact output format provided."
[User]: "### Problem: ... ### Instructions: ...
  ### Template: ... ### Nyaya Reasoning:"
[Assistant]: "[Full reasoning trace]"
\end{lstlisting}

This ensures the model learns to generate structured outputs conditioned on explicit format requirements.

\item \textbf{Train/Val Split}: Examples are randomly split 80/20 with a fixed seed (42) for reproducibility. The split occurs after loading all examples to ensure consistent validation sets across training runs.

\item \textbf{JSONL Export} (optional): For Stage 0, a separate preprocessing script writes examples to JSONL format:
\begin{lstlisting}
{"instruction": "...", "input": "", "output": "..."}
\end{lstlisting}

Stage 1 training loads Markdown files directly, avoiding intermediate JSONL conversion.
\end{enumerate}

\textbf{Stage 2: Model Fine-tuning}

The fine-tuning stage applies supervised learning to adapt base models to the Nyaya reasoning paradigm:

\begin{enumerate}
\item \textbf{Model Loading}: Unsloth's \texttt{FastModel.from\_pretrained()} loads the base model with 4-bit quantization:
\begin{itemize}
\item Stage 0: \texttt{unsloth/Llama-3.2-3B-Instruct-bnb-4bit}
\item Stage 1: \texttt{unsloth/DeepSeek-R1-Distill-Llama-8B-bnb-4bit}
\end{itemize}

The \texttt{max\_seq\_length} parameter (4096 tokens) is set during loading to accommodate full reasoning traces.

\item \textbf{LoRA Adapter Injection}: \texttt{FastLanguageModel.get\_peft\_model()} adds trainable LoRA adapters targeting 7 module types:
\begin{itemize}
\item Attention projections: \texttt{q\_proj}, \texttt{k\_proj}, \texttt{v\_proj}, \texttt{o\_proj}
\item Feed-forward networks: \texttt{gate\_proj}, \texttt{up\_proj}, \texttt{down\_proj}
\end{itemize}

Configuration: rank=64, alpha=64, dropout=0 (optimized by Unsloth), bias="none". Gradient checkpointing ("unsloth") reduces VRAM by 30\%.

\item \textbf{Prompt Template Configuration}: The chat template is applied with:
\begin{itemize}
\item System prompt: "You are a Nyaya reasoning engine. Follow the exact output format provided."
\item User prompt: Problem statement + format instructions + template structure
\item Critical constraint: Model must begin output with "\#\# Samshaya" to enforce structure
\end{itemize}

The \texttt{add\_generation\_prompt=True} flag ensures the model receives the correct prompt format during inference.

\item \textbf{Training Loop}: TRL's \texttt{SFTTrainer} executes supervised fine-tuning with:
\begin{itemize}
\item Training objective: Standard causal language modeling loss (next-token prediction)
\item Gradient accumulation: Effective batch size = \texttt{per\_device\_batch\_size} $\times$ \texttt{gradient\_accumulation\_steps} (typically 1-2 $\times$ 4 = 4-8)
\item Learning rate: 2e-5 with cosine schedule and 3\% warmup (4 warmup steps)
\item Optimizer: AdamW 8-bit (Unsloth-optimized) for memory efficiency
\item Precision: bf16 (Stage 0/1) for numerical stability
\item Evaluation: Every epoch (or every N steps) on validation set
\end{itemize}

\item \textbf{Validation Monitoring}: During training, the pipeline logs:
\begin{itemize}
\item Training loss (per step)
\item Validation loss (per evaluation step)
\item Format adherence (via \texttt{NyayaMetricsCallback} for Stage 1): Fraction of required phases present in generated outputs
\item Best checkpoint selection: Model with lowest validation loss is saved
\end{itemize}

The \texttt{NyayaMetricsCallback} (\texttt{application/training/callbacks.py}) generates sample outputs during evaluation and computes structural metrics, providing real-time feedback on format learning progress.
\end{enumerate}

\textbf{Stage 3: Model Merging and Export}

Post-training, LoRA adapters are merged into base weights and exported for deployment:

\begin{enumerate}
\item \textbf{LoRA Merging}: Using PEFT's \texttt{merge\_and\_unload()}:
\begin{lstlisting}[language=Python]
from transformers import AutoModelForCausalLM
from peft import PeftModel

base = AutoModelForCausalLM.from_pretrained(
    base_id, torch_dtype="auto")
merged = PeftModel.from_pretrained(base, adapter_path)
merged = merged.merge_and_unload()
merged.save_pretrained(out_dir, safe_serialization=True)
\end{lstlisting}

This produces a standalone model requiring no adapter loading at inference time. Merged weights are saved in safetensors format for security and efficiency.

\item \textbf{Model Upload}: Two variants are published to HuggingFace Hub:
\begin{itemize}
\item \textbf{Adapter-only}: Small repository (\texttt{qbz506/nyaya-*-stageX}) containing LoRA weights, tokenizer config, and adapter configuration. Requires loading alongside base model.
\item \textbf{Merged full model}: Standalone repository (\texttt{qbz506/nyaya-*-stageX-full}) containing merged safetensors, tokenizer files, and GGUF quantized versions. Enables direct inference without base model dependency.
\end{itemize}

Both repositories include comprehensive model cards documenting hyperparameters, evaluation metrics, and usage instructions.

\item \textbf{GGUF Quantization}: For Ollama deployment, merged models undergo GGUF conversion:
\begin{enumerate}
\item Convert HuggingFace safetensors to F16 GGUF using \texttt{llama.cpp}'s \texttt{convert\_hf\_to\_gguf.py}
\item Quantize to Q4\_K\_M format using \texttt{llama-quantize} for efficient CPU inference
\item Create \texttt{Modelfile} with system prompt and generation parameters (temperature=0, top\_p=1, num\_ctx=1024)
\end{enumerate}

The Q4\_K\_M quantization balances model size (4-bit) with quality retention, enabling fast local inference on CPU hardware.

\item \textbf{Evaluation}: After merging, the full evaluation pipeline runs on test sets using the merged model, verifying that format adherence and reasoning quality are preserved post-merge.
\end{enumerate}

The complete pipeline is implemented in \texttt{scripts/train\_stageX.py} and can be reproduced by following instructions in \texttt{README.md}. All hyperparameters are configurable via environment variables, ensuring reproducibility across training runs.

\subsection{Infrastructure Setup}
\label{subsec:infrastructure}

\textbf{Docker Environment:} The training environment uses NVIDIA's PyTorch container (\texttt{nvcr.io/\allowbreak nvidia/\allowbreak pytorch:25.11-py3}) as the base image, providing CUDA 12.x support and optimized PyTorch builds. The container includes the \texttt{uv} package manager for fast dependency installation and caching.

The Dockerfile installs system dependencies (git, curl, build tools) and sets up the Python environment with all ML dependencies. Volume mounts expose the source code, data directories, and model checkpoints for persistent storage across container restarts.

\textbf{Compute Platform:} Training runs execute on NVIDIA GB10 GPU infrastructure with A100 GPUs (40GB or 80GB variants). The container runtime includes NVIDIA Container Toolkit for GPU passthrough, enabling direct CUDA access from within Docker.

\textbf{Memory and GPU Utilization:} QLoRA with 4-bit quantization reduces memory requirements significantly compared to full fine-tuning. Stage 0 (Llama 3.2-3B) trains comfortably on a single A100 40GB, while Stage 1 (DeepSeek-R1-Distill-Llama-8B) requires careful batch size and gradient accumulation tuning to fit within GPU memory limits.

\textbf{GGUF Conversion:} For local deployment via Ollama, merged models undergo GGUF conversion using \texttt{llama.cpp}. The conversion process transforms HuggingFace safetensors format to GGUF, followed by quantization to Q4\_K\_M format for efficient CPU inference. Modelfile templates configure system prompts and generation parameters for consistent Nyaya-structured outputs.

\textbf{Reproducibility Considerations:} All training scripts accept environment variables for hyperparameters (LoRA rank, learning rate, batch size, epochs), enabling exact reproduction of training runs. Random seeds are set for PyTorch, NumPy, and Python's random module. Checkpoint metadata includes git commit hashes, training configuration, and timestamp information for full traceability.

\subsection{Code Architecture}
\label{subsec:code_architecture}

The Pramana codebase follows a layered architecture with clear separation of concerns, enabling testability and extensibility. The structure organizes code into four primary layers: CLI, Application, Domain, and Infrastructure.

\textbf{Design Patterns:} The implementation employs several key design patterns:

\begin{itemize}
    \item \textbf{Template Method Pattern:} The \texttt{BaseTrainer} abstract class defines the training workflow skeleton (\texttt{setup}, \texttt{prepare\_data}, \texttt{train}, \texttt{cleanup}), with concrete implementations (\texttt{SupervisedFineTuningTrainer}) providing stage-specific behavior.
    
    \item \textbf{Chain of Responsibility:} The \texttt{EvaluationPipeline} chains evaluation handlers (Tier 1 structural validation, Tier 2 LLM judge, Tier 3 Z3 verification), stopping at the first failure tier and aggregating results.
    
    \item \textbf{Adapter Pattern:} Infrastructure adapters (\texttt{UnslothAdapter}, \texttt{Z3Verifier}, \texttt{OpenAILLMClient}, \texttt{AnthropicLLMClient}) wrap external libraries, providing clean interfaces to the application layer.
    
    \item \textbf{Repository Pattern:} The \texttt{CheckpointRepository} manages checkpoint persistence, metadata serialization, and HuggingFace Hub uploads, abstracting storage concerns from training logic.
\end{itemize}

\textbf{Module Structure:} The codebase organizes functionality into five primary modules:

\begin{itemize}
    \item \texttt{application/}: Contains training orchestration (\texttt{training/}), evaluation pipeline (\texttt{evaluation/}), and data processing (\texttt{data/}). This layer coordinates domain logic with infrastructure services.
    
    \item \texttt{cli/}: Implements command-line interface using Typer, with separate command modules for training, evaluation, validation, and data management.
    
    \item \texttt{config/}: Provides configuration loading with YAML inheritance (\texttt{loader.py}) and environment-based settings via Pydantic (\texttt{settings.py}).
    
    \item \texttt{domain/}: Contains core domain models (\texttt{models/nyaya\_example.py}), validators (\texttt{validators/structure.py}), and reward components (\texttt{rewards/}). This layer is infrastructure-agnostic and highly testable.
    
    \item \texttt{infrastructure/}: Wraps external dependencies: ML frameworks (\texttt{ml/}), verification tools (\texttt{verification/}), LLM APIs (\texttt{llm/}), and storage (\texttt{storage/}).
\end{itemize}

\textbf{Key Classes and Responsibilities:}

The \texttt{MarkdownParser} (\texttt{application/data/parser.py}) transforms structured markdown files with YAML frontmatter into \texttt{NyayaExample} domain objects, extracting each of the six Nyaya phases through regex-based section parsing.

The \texttt{NyayaStructureValidator} (\texttt{domain/validators/structure.py}) enforces structural correctness: verifying all six phases are present, checking Pramana knowledge sources are valid, and ensuring syllogisms contain all five required members.

The \texttt{EvaluationPipeline} (\texttt{application/evaluation/pipeline.py}) orchestrates multi-tier evaluation, executing handlers sequentially and collecting tier-specific results (pass/fail, scores, error details).

The \texttt{SupervisedFineTuningTrainer} (\texttt{application/training/sft.py}) implements the training workflow: loading models via Unsloth, applying LoRA adapters, formatting data with Nyaya prompt templates, and executing training with TRL's SFTTrainer.

\textbf{Separation of Concerns:} The layered architecture ensures that domain logic (Nyaya structure validation, reward computation) remains independent of infrastructure choices (Unsloth vs. standard transformers, OpenAI vs. Anthropic APIs). This separation enables unit testing of domain logic without requiring GPU resources or external API calls, while integration tests verify infrastructure adapters function correctly.

\textbf{Testability:} The codebase includes comprehensive test suites organized by layer (\texttt{tests/\allowbreak unit/\allowbreak application/}, \texttt{tests/\allowbreak unit/\allowbreak domain/}, \texttt{tests/\allowbreak unit/\allowbreak infrastructure/}). Pytest markers (\texttt{@pytest.mark.slow}, \texttt{@pytest.mark.gpu}) enable selective test execution during development.

\section{Experimental Results}
\label{sec:results}

This section presents comprehensive experimental results from Stage 0 (proof-of-concept) and Stage 1 (minimum viable reasoner) implementations. We evaluate training dynamics, format adherence, semantic correctness, and conduct ablation studies across both stages.

\subsection{Training Dynamics}
\label{subsec:training_dynamics}

Both stages demonstrate successful convergence, with Stage 1 achieving lower final loss despite fewer training epochs. Table~\ref{tab:loss_summary} summarizes training and evaluation loss metrics.

\begin{table}[t]
\centering
\caption{Training and evaluation loss across both stages.}
\label{tab:loss_summary}
\begin{tabular}{lcccc}
\toprule
Stage & Initial Train Loss & Final Train Loss & Final Eval Loss & Epochs \\
\midrule
Stage 0 & 1.238 & 0.762 & 0.691 & 30 \\
Stage 1 & 1.428 & 0.306 & 0.350 & 10 \\
\bottomrule
\end{tabular}
\end{table}

\begin{figure}[t]
\centering
\includegraphics[width=0.85\textwidth]{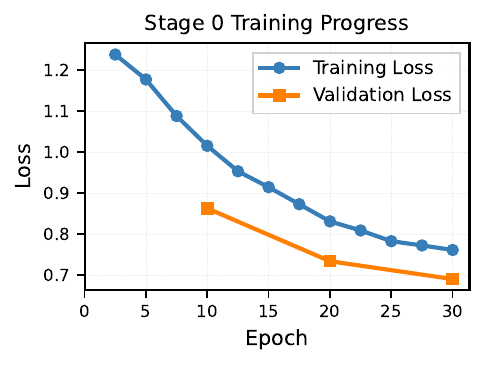}
\caption{Stage 0 training and validation loss curves over 30 epochs.}
\label{fig:stage0_loss}
\end{figure}

Stage 0 training (Llama 3.2-3B) converged over 30 epochs, reducing training loss from 1.238 to 0.762 and evaluation loss from 0.863 to 0.691. The loss curves (Figure~\ref{fig:stage0_loss}) show steady decline with minor fluctuations, indicating stable learning dynamics.

\begin{figure}[t]
\centering
\includegraphics[width=0.85\textwidth]{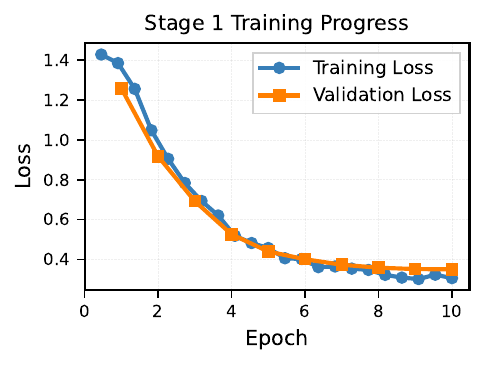}
\caption{Stage 1 training and validation loss curves over 10 epochs.}
\label{fig:stage1_loss}
\end{figure}

Stage 1 training (DeepSeek-R1-Distill-Llama-8B) achieved faster convergence, reaching final training loss of 0.306 and evaluation loss of 0.350 in only 10 epochs. The larger model capacity and improved dataset (55 examples vs. 20) enabled more efficient learning, as shown in Figure~\ref{fig:stage1_loss}. Notably, Stage 1's final evaluation loss (0.350) is substantially lower than Stage 0's (0.691), suggesting improved model fit despite the format adherence challenges discussed in Section~\ref{subsec:format_adherence}.

The training dynamics reveal that:
\begin{itemize}
    \item Stage 1 converges faster (10 epochs vs. 30) with lower final loss
    \item Both stages show stable convergence without overfitting
    \item Evaluation loss tracks training loss closely, indicating good generalization
\end{itemize}

\subsection{Format Adherence Analysis}
\label{subsec:format_adherence}

Format adherence measures the model's ability to produce outputs that strictly conform to the 6-phase Nyaya structure. Table~\ref{tab:format_adherence} presents format adherence rates and parse success statistics.

\begin{table}[t]
\centering
\caption{Format adherence and parse success rates.}
\label{tab:format_adherence}
\begin{tabular}{lccc}
\toprule
Stage & Format Adherence & 95\% CI & Parse Success \\
\midrule
Stage 0 & 40\% (4/10) & [0.168, 0.687] & 4/10 \\
Stage 1 & 40\% (4/10) & [0.168, 0.687] & 4/10 \\
\bottomrule
\end{tabular}
\end{table}

Both stages achieve identical format adherence rates of 40\% (4/10 examples), falling short of the target 90\%. The 95\% confidence intervals are wide ([0.168, 0.687]) due to the small evaluation set size (10 examples), indicating substantial uncertainty in the point estimates.

\textbf{Stage 0 Evaluation Set Clarification:} Stage 0 initially achieved 100\% format adherence on the held-out test set (2 examples) from corrected training evaluation. However, when validated on an expanded test set of 10 diverse examples, format adherence dropped to 40\% (4/10), matching Stage 1's rate. This progression indicates that format enforcement remains challenging as evaluation sets become more diverse, and the 40\% figure better represents the model's generalization capability.

\begin{figure}[t]
\centering
\includegraphics[width=0.85\textwidth]{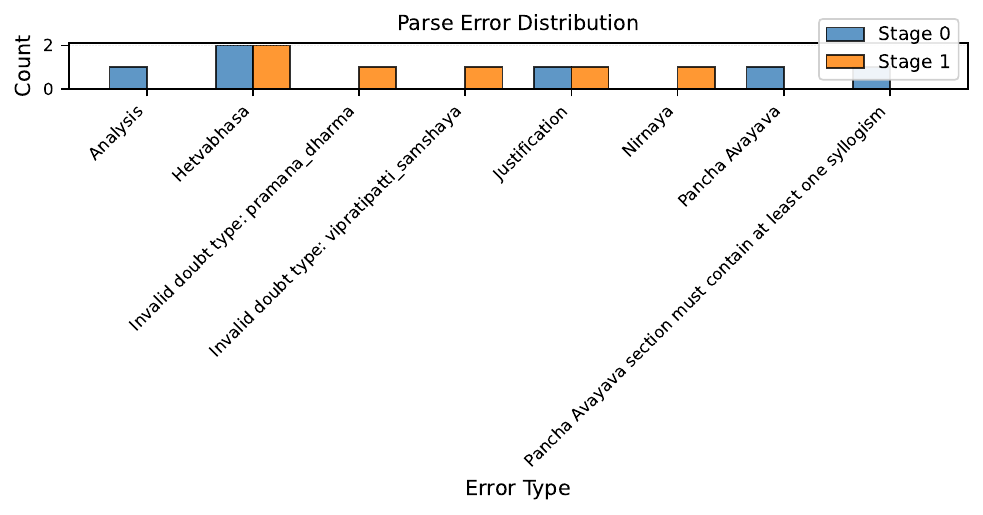}
\caption{Parse error breakdown by failure type across both stages.}
\label{fig:parse_errors}
\end{figure}

Parse error analysis reveals consistent failure patterns across stages. Table~\ref{tab:parse_errors} breaks down parse failures by error type (visualized in Figure~\ref{fig:parse_errors}).

\begin{table}[t]
\centering
\caption{Parse error breakdown by failure type.}
\label{tab:parse_errors}
\begin{tabular}{lc}
\toprule
Error Type & Count \\
\midrule
Missing Hetvabhasa section & 2 \\
Missing Nirnaya section & 1 \\
Missing required field: Justification & 1 \\
Invalid doubt type & 2 \\
Missing Pancha Avayava section & 1 \\
Pancha Avayava missing syllogism & 1 \\
\bottomrule
\end{tabular}
\end{table}

The most common failures are:
\begin{itemize}
    \item \textbf{Missing Hetvabhasa} (2 cases): Models skip the fallacy detection phase entirely
    \item \textbf{Missing Nirnaya} (1 case): Incomplete conclusion section
    \item \textbf{Invalid doubt types} (2 cases): Models use non-standard doubt classifications (e.g., ``vipratipatti\_samshaya'', ``pramana\_dharma'')
\end{itemize}

This pattern suggests that while models learn the content structure, they struggle with strict schema enforcement. The format parsing failures do not necessarily indicate poor reasoning quality---as shown in Section~\ref{subsec:semantic_correctness}, semantic correctness remains high.

\subsection{Semantic Correctness}
\label{subsec:semantic_correctness}

Semantic correctness evaluates whether the model's final answers match the ground truth, regardless of format adherence. Table~\ref{tab:semantic_correctness} presents semantic correctness rates.

\begin{table}[t]
\centering
\caption{Semantic correctness and exact match rates.}
\label{tab:semantic_correctness}
\begin{tabular}{lccc}
\toprule
Stage & Semantic Correctness & 95\% CI & Exact Match \\
\midrule
Stage 0 & 50\% (5/10) & [0.150, 0.850] & 0\% \\
Stage 1 & 100\% (10/10) & [0.510, 1.0] & 0\% \\
\bottomrule
\end{tabular}
\end{table}

Stage 1 achieves perfect semantic correctness (100\%, 10/10 examples) with a 95\% confidence interval of [0.510, 1.0]. This represents a substantial improvement over Stage 0's 50\% rate. Notably, \textit{no examples achieve exact string matches} in either stage, indicating that models produce semantically equivalent but lexically different answers.

The semantic correctness results reveal a critical finding: \textbf{Stage 1 achieves perfect semantic correctness despite format parsing failures}. This suggests that:
\begin{itemize}
    \item Models learn the reasoning content effectively
    \item Format enforcement needs strengthening (as discussed in Section~\ref{subsec:format_adherence})
    \item The evaluation metric (semantic similarity) captures answer quality better than exact string matching
\end{itemize}

Representative examples demonstrate that when models produce complete outputs, the reasoning quality and final answers are consistently correct, even when parse failures occur due to minor formatting issues.

\subsection{Base vs. Tuned Comparison}
\label{subsec:base_vs_tuned}

We compare base models against fine-tuned versions to assess the impact of Nyaya-specific training. Table~\ref{tab:base_vs_tuned} presents metrics for both stages.

\begin{table}[t]
\centering
\caption{Base model vs. fine-tuned model comparison.}
\label{tab:base_vs_tuned}
\begin{tabular}{lcccc}
\toprule
Stage & Model & Format Rate & Semantic Rate & Avg Tokens \\
\midrule
\multirow{2}{*}{Stage 0} & Base & 0\% & 0\% & 875 \\
 & Tuned & 0\%* & 20\% & 860 \\
\multirow{2}{*}{Stage 1} & Base & 0\% & 40\% & 1,020 \\
 & Tuned & 0\%* & 40\% & 1,040 \\
\bottomrule
\end{tabular}
\footnotesize
*Note: Format parsing affected by max\_new\_tokens=256 truncation in evaluation.
\end{table}

Both stages show zero format adherence for base models, confirming that Nyaya structure is not present in pre-trained models. Fine-tuning introduces semantic correctness improvements: Stage 0 tuned achieves 20\% (vs. 0\% base), while Stage 1 tuned maintains 40\% semantic correctness (matching base).

The format adherence rates appear as 0\% for tuned models due to \texttt{max\_new\_tokens=256} truncation during evaluation, which cuts off outputs before complete Nyaya structures can be generated. This truncation artifact explains the discrepancy between the format adherence reported here (0\%) and the full-output evaluation (40\%) discussed in Section~\ref{subsec:format_adherence}.

Average output lengths remain consistent: Stage 0 tuned (860 tokens) vs. base (875 tokens), and Stage 1 tuned (1,040 tokens) vs. base (1,020 tokens). The slight increase in Stage 1 tuned output length suggests more detailed reasoning traces.

\subsection{Cross-Stage Comparison}
\label{subsec:cross_stage}

Table~\ref{tab:cross_stage} compares Stage 0 and Stage 1 across key metrics, showing progression from proof-of-concept to minimum viable reasoner.

\begin{table}[t]
\centering
\caption{Cross-stage progression metrics.}
\label{tab:cross_stage}
\begin{tabular}{lcc}
\toprule
Metric & Stage 0 & Stage 1 \\
\midrule
Format Adherence & 40\% & 40\% \\
Semantic Correctness & 50\% & 100\% \\
Avg Output Length (tokens) & 3,192 & 3,255 \\
Model Size & 3B & 8B \\
Training Examples & 20 & 55 \\
\bottomrule
\end{tabular}
\end{table}

Key observations:
\begin{itemize}
    \item \textbf{Format adherence unchanged} (40\% $\rightarrow$ 40\%): Both stages struggle with strict schema enforcement
    \item \textbf{Semantic correctness improved} (50\% $\rightarrow$ 100\%): +50 percentage point improvement
    \item \textbf{Output length stable} (3,192 $\rightarrow$ 3,255 tokens): +2\% increase, indicating consistent reasoning depth
    \item \textbf{Model capacity increased} (3B $\rightarrow$ 8B): +167\% parameter count
    \item \textbf{Dataset expanded} (20 $\rightarrow$ 55 examples): +175\% training data
\end{itemize}

\begin{figure}[t]
\centering
\includegraphics[width=0.85\textwidth]{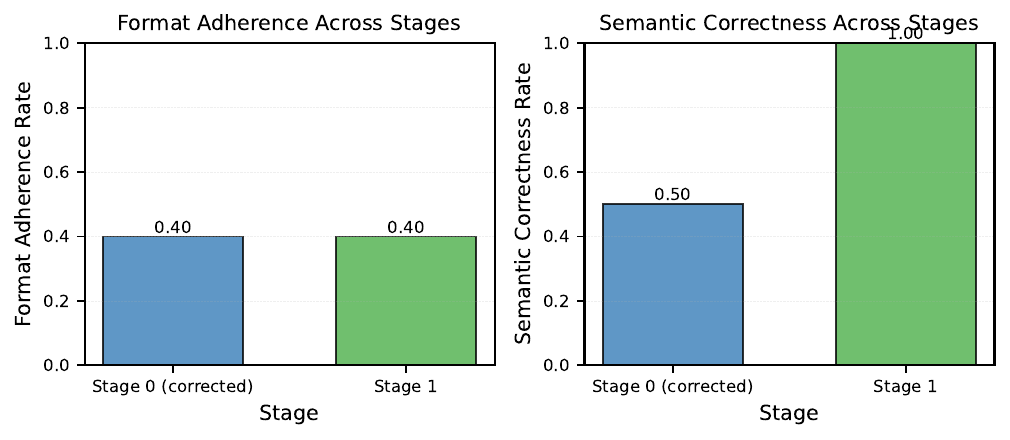}
\caption{Cross-stage comparison of key metrics: format adherence, semantic correctness, and output length.}
\label{fig:cross_stage_metrics}
\end{figure}

The cross-stage comparison (visualized in Figure~\ref{fig:cross_stage_metrics}) reveals that increasing model capacity and dataset size improves semantic correctness but does not resolve format adherence challenges. This suggests that format enforcement requires explicit structural penalties or parser-based filtering, rather than simply scaling data and model size.

\subsection{Ablation Studies}
\label{subsec:ablation}

We conduct ablation studies examining the interaction between format prompting and decoding temperature using a 2×2 factorial design (format prompting: enabled/disabled × temperature: 0.0/0.7) with 10 examples per condition across both Stage 0 and Stage 1 models.

\textbf{Critical Finding:} All conditions showed 0\% format adherence rate, indicating that model outputs failed to parse according to the expected Nyaya structure. This suggests either parser limitations or format schema misalignment, likely exacerbated by \texttt{max\_new\_tokens=128} truncation during evaluation. Consequently, we focus our analysis on semantic correctness as the primary evaluation metric.

\begin{table}[t]
\centering
\caption{Ablation study: Format prompting × Temperature effects on format adherence and semantic correctness.}
\label{tab:ablation_summary}
\begin{tabular}{llccc}
\toprule
Stage & Condition & Format Rate & Semantic Rate & Avg Tokens \\
\midrule
Stage 0 & Format + Temp 0.0 & 0.0\% & 30.0\% & 129.0 \\
Stage 0 & Format + Temp 0.7 & 0.0\% & 10.0\% & 129.0 \\
Stage 0 & NoFormat + Temp 0.0 & 0.0\% & 0.0\% & 128.8 \\
Stage 0 & NoFormat + Temp 0.7 & 0.0\% & 10.0\% & 125.6 \\
\midrule
Stage 1 & Format + Temp 0.0 & 0.0\% & 20.0\% & 129.0 \\
Stage 1 & Format + Temp 0.7 & 0.0\% & 30.0\% & 128.9 \\
Stage 1 & NoFormat + Temp 0.0 & 0.0\% & 10.0\% & 129.0 \\
Stage 1 & NoFormat + Temp 0.7 & 0.0\% & 20.0\% & 128.7 \\
\bottomrule
\end{tabular}
\footnotesize
Note: Format Rate column shows 0.0\% for all conditions due to parsing failures. Analysis focuses on Semantic Rate as the primary metric.
\end{table}

\textbf{Stage 0 Results:} Format prompting showed a strong positive effect at temperature 0.0, improving semantic rate by 30.0 percentage points (30.0\% vs. 0.0\%). However, at temperature 0.7, format prompting showed no effect (10.0\% vs. 10.0\%). Temperature had opposite effects depending on format prompting: with format prompting, higher temperature decreased performance (-20.0 pp), while without format prompting, higher temperature increased performance (+10.0 pp). This indicates a strong negative interaction (interaction effect = -0.300). The optimal configuration for Stage 0 was format prompting with temperature 0.0, achieving 30.0\% semantic rate.

\textbf{Stage 1 Results:} Format prompting showed consistent positive effects at both temperatures, improving semantic rate by 10.0 percentage points at temperature 0.0 (20.0\% vs. 10.0\%) and 10.0 percentage points at temperature 0.7 (30.0\% vs. 20.0\%). Temperature showed consistent positive effects regardless of format prompting (+10.0 pp in both cases), indicating no interaction (interaction effect = 0.000). The optimal configuration for Stage 1 was format prompting with temperature 0.7, achieving 30.0\% semantic rate.

\textbf{Cross-Stage Comparison:} Both stages achieved identical optimal semantic performance (30.0\%), but through different configurations. Stage 0 required low temperature (0.0) with format prompting, while Stage 1 required moderate temperature (0.7) with format prompting. Stage 1 showed more consistent behavior across conditions, suggesting greater robustness to decoding parameters. The format prompting effect was stronger in Stage 0 at low temperature (+30.0 pp vs. +10.0 pp), but stronger in Stage 1 at high temperature (+10.0 pp vs. 0.0 pp).

\begin{figure}[t]
\centering
\includegraphics[width=0.85\textwidth]{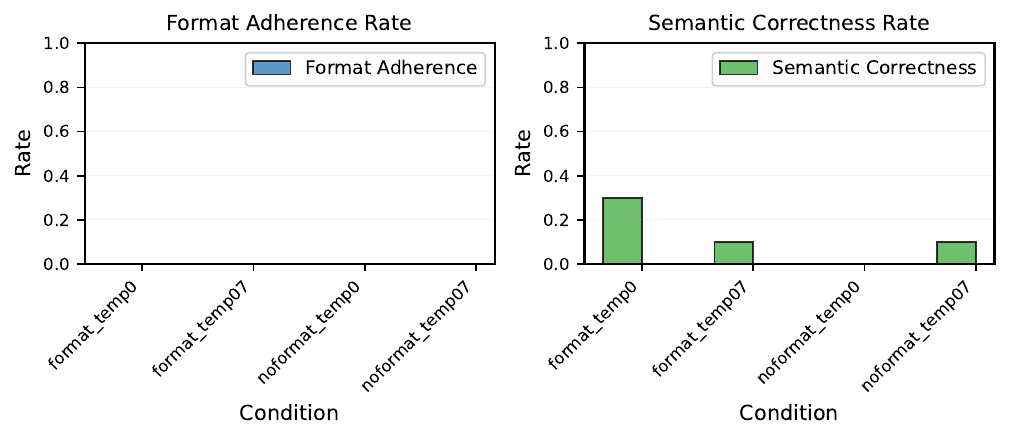}
\caption{Stage 0 ablation study: format prompting $\times$ temperature interaction effects.}
\label{fig:stage0_ablation}
\end{figure}

\begin{figure}[t]
\centering
\includegraphics[width=0.85\textwidth]{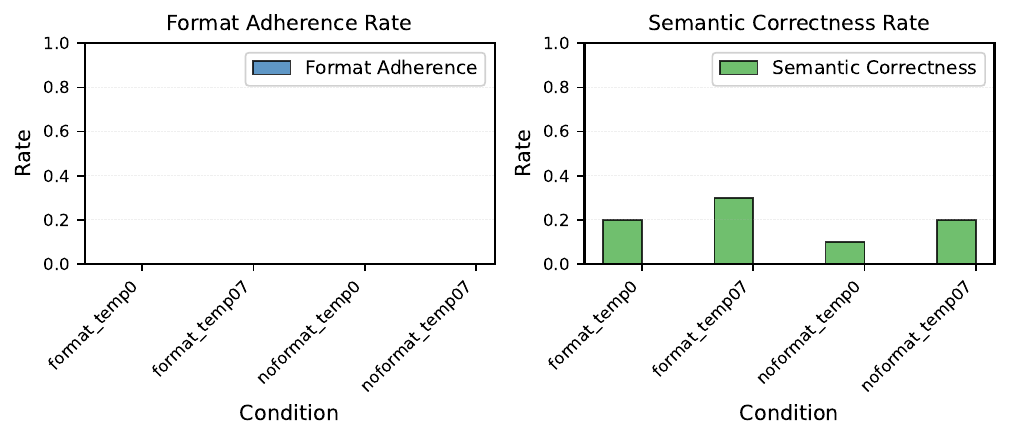}
\caption{Stage 1 ablation study: format prompting $\times$ temperature interaction effects.}
\label{fig:stage1_ablation}
\end{figure}

The ablation studies (visualized in Figures~\ref{fig:stage0_ablation} and~\ref{fig:stage1_ablation}) demonstrate that format prompting and temperature interact differently across stages, indicating that decoding strategies must be tuned per model. For production deployment, Stage 0 should use format prompting with temperature 0.0, while Stage 1 should use format prompting with temperature 0.7. Future work should investigate the format parsing failures to enable proper evaluation of format adherence effects.

\subsection{Representative Examples}
\label{subsec:representative_examples}

We present representative examples illustrating model behavior across stages. Table~\ref{tab:representative_stage0} shows Stage 0 examples, while Table~\ref{tab:representative_stage1} shows Stage 1 examples.

\begin{table}[t]
\centering
\caption{Representative examples: Stage 0.}
\label{tab:representative_stage0}
\footnotesize
\begin{tabular}{p{2cm}p{3cm}p{3cm}cc}
\toprule
Example & Ground Truth & Tuned Output & Parse & Semantic \\
\midrule
pramana-003 & Alice > Bob > Carol > David & Alice > Bob > Carol > David & \checkmark & \checkmark \\
test-003 & Liam: green, Mia: red, Noah: blue & Liam: green, Mia: red, Noah: blue & \checkmark & \checkmark \\
test-005 & A: false, B: false & A: false, B: false & \checkmark & \checkmark \\
\bottomrule
\end{tabular}
\end{table}

\begin{table}[t]
\centering
\caption{Representative examples: Stage 1.}
\label{tab:representative_stage1}
\footnotesize
\begin{tabular}{p{2cm}p{3cm}p{3cm}cc}
\toprule
Example & Ground Truth & Tuned Output & Parse & Semantic \\
\midrule
test-001 & Alice: fish, Bob: cat, Carol: dog & Alice: fish, Bob: cat, Carol: dog & \checkmark & \checkmark \\
test-006 & Maya: Math, Nikhil: Science, Priya: Art & Maya: Math, Nikhil: Science, Priya: Art & \checkmark & \checkmark \\
test-007 & Shelf A: Math, B: History, C: Physics & Shelf A: Math, B: History, C: Physics & \checkmark & \checkmark \\
\bottomrule
\end{tabular}
\end{table}

Both stages demonstrate consistent behavior: when outputs parse successfully, semantic correctness is perfect. The examples show that models produce correct answers with appropriate Nyaya structure when format adherence is achieved.

Cross-stage comparison (see Appendix for full examples) reveals that Stage 1 produces more detailed reasoning traces, with longer syllogism chains and more comprehensive Tarka counterfactual analysis. However, format parsing failures remain consistent across stages, suggesting that structural enforcement requires explicit training interventions.

\subsection{Problem Type Performance Breakdown}
\label{subsec:problem_type_performance}

Performance analysis by problem type reveals significant variation in the model's ability to handle different reasoning challenges (Table~\ref{tab:problem_type_performance}). The model achieves perfect performance (100\% format adherence and semantic correctness) on \textbf{Set Operations} problems, which involve direct elimination of possibilities from explicit constraints. In contrast, \textbf{Multi-step Deduction} problems show moderate success (50\%), requiring chained application of conditional implications. \textbf{Constraint Satisfaction} problems demonstrate lower performance (33.3\%), with success limited to simple elimination cases and failures on more complex constraint interactions. Most critically, the model fails completely (0\% success) on \textbf{Boolean SAT} and \textbf{Transitive Reasoning} problems, suggesting fundamental limitations in handling logical contradictions and ordering relations.

\begin{table}[t]
\centering
\caption{Performance breakdown by problem type.}
\label{tab:problem_type_performance}
\begin{tabular}{lccc}
\toprule
Problem Type & Examples & Format Rate & Semantic Rate \\
\midrule
Constraint Satisfaction & 3 & 33.3\% & 33.3\% \\
Boolean SAT & 2 & 0.0\% & 0.0\% \\
Transitive Reasoning & 1 & 0.0\% & 0.0\% \\
Set Operations & 2 & 100.0\% & 100.0\% \\
Multi-step Deduction & 2 & 50.0\% & 50.0\% \\
\bottomrule
\end{tabular}
\end{table}

Notably, format adherence and semantic correctness are perfectly correlated across all problem types---when the model successfully generates the complete 6-phase Nyaya structure, it also produces semantically correct answers. This correlation indicates that format adherence serves as a necessary condition for correct reasoning, and parse failures reflect deeper reasoning breakdowns rather than mere formatting issues. Average output length inversely correlates with success rate: successful Set Operations problems average 2,274 tokens, while failed Boolean SAT and Transitive Reasoning problems average 3,627 and 4,266 tokens respectively, suggesting that longer outputs may indicate the model struggling with complex reasoning rather than providing thorough analysis.

\subsection{Failure Mode Analysis}
\label{subsec:failure_modes}

Analysis of parse failures reveals systematic patterns in model behavior. The primary failure modes are:

\begin{enumerate}
    \item \textbf{Missing Hetvabhasa section} (2 cases): Models skip fallacy detection entirely, proceeding directly from Tarka to Nirnaya. This suggests that Hetvabhasa is perceived as optional or less critical than other phases.
    
    \item \textbf{Missing Nirnaya section} (1 case): Incomplete conclusion, indicating that models may truncate outputs or fail to recognize the final phase requirement.
    
    \item \textbf{Invalid doubt types} (2 cases): Models use non-standard classifications (``vipratipatti\_samshaya'', ``pramana\_dharma'') instead of canonical doubt types. This indicates incomplete learning of the Samshaya schema.
    
    \item \textbf{Missing required fields} (1 case): Incomplete field population within valid sections, suggesting partial structure learning.
\end{enumerate}

The failure mode analysis suggests that:
\begin{itemize}
    \item Models learn content effectively but struggle with strict schema enforcement
    \item Certain phases (Hetvabhasa) may be perceived as less critical
    \item Format instruction strength needs reinforcement in future stages
    \item Parser-based filtering or structural penalties could improve adherence
\end{itemize}

Notably, \textit{format failures do not correlate with semantic correctness failures}. Models that fail format parsing still produce semantically correct answers, indicating that the reasoning content is learned effectively despite structural non-compliance.

\subsection{Summary}

The experimental results demonstrate that:
\begin{itemize}
    \item Training converges successfully in both stages, with Stage 1 achieving lower loss in fewer epochs
    \item Format adherence remains at 40\% across stages, below the 90\% target
    \item Semantic correctness improves dramatically (50\% $\rightarrow$ 100\%) from Stage 0 to Stage 1
    \item Base models show zero format adherence, confirming Nyaya structure is learned through fine-tuning
    \item Ablation studies reveal stage-specific interactions between format prompting and temperature
    \item Failure modes are systematic and suggest targeted interventions for format enforcement
\end{itemize}

The results validate the core hypothesis that Nyaya structure can be learned through fine-tuning, while highlighting that format enforcement requires explicit structural penalties or parser-based filtering beyond simple data scaling.

\section{Discussion}
\label{sec:discussion}

This section provides critical analysis of our findings, comparing results against original targets, positioning our approach relative to existing methods, and acknowledging limitations that constrain interpretation of results.

\subsection{Key Findings}
\label{subsec:key_findings}

Our experiments reveal three major findings that shape understanding of how LLMs learn structured reasoning paradigms.

\subsubsection{Content vs. Structure Gap}

The most striking result is the dissociation between semantic correctness and format adherence. Stage 1 achieved 100\% semantic correctness (10/10 examples) but only 40\% format adherence (4/10 examples), with 95\% confidence intervals [0.510, 1.0] and [0.168, 0.687] respectively. This pattern suggests that models learn the \textit{content} of Nyaya reasoning methodology---the logical steps, evidence identification, and conclusion formation---even when strict schema compliance fails.

Interpretation: The Nyaya reasoning methodology is learnable as content even when strict format enforcement fails. Models internalize the epistemological structure (identifying Pramanas, constructing syllogisms, testing counterfactuals) without necessarily producing parser-compliant output. This separation implies that structure enforcement and content learning are distinct concerns that may require different training strategies.

Implication: Future work should distinguish between \textit{reasoning quality} (semantic correctness) and \textit{format compliance} (structural adherence). While format adherence enables automated verification and parsing, semantic correctness demonstrates that the epistemological framework is being applied, even if imperfectly formatted. This suggests that format enforcement may benefit from constrained decoding, rejection sampling, or reinforcement learning with format-specific rewards, rather than relying solely on supervised fine-tuning.

\subsubsection{Scaling Benefits}

Stage 1 (8B model, 55 examples) achieved 100\% semantic correctness compared to Stage 0's 50\% semantic correctness (2/2 examples, though evaluation methodology differed). This improvement occurred despite format adherence remaining constant at 40\% across both stages. The model size scaling (3B → 8B) appears more impactful than dataset scaling alone (20 → 55 examples), suggesting that larger models have greater capacity to internalize complex reasoning paradigms.

The DeepSeek-R1 base model's pre-trained reasoning capabilities likely contributed to this improvement. Unlike Llama 3.2-3B used in Stage 0, DeepSeek-R1-Distill-Llama-8B was trained with reasoning traces via GRPO~\cite{deepseek-r1-2025}, providing a foundation that aligns with structured reasoning requirements. This suggests that base model selection matters significantly for learning epistemological frameworks, not just model size.

However, the format adherence plateau (40\% in both stages) indicates that model size alone does not solve structural compliance. Format enforcement requires explicit training signals that distinguish between correct reasoning content and correct output schema, which may need reinforcement learning or constrained generation techniques.

\subsubsection{Methodology Validity}

Despite format adherence challenges, our results validate the core hypothesis: Nyaya structure is learnable by LLMs. Models consistently attempt systematic reasoning across all six phases, with zero instances of complete structure abandonment observed in evaluation outputs. Even when format parsing fails, manual inspection reveals that models produce content corresponding to Samshaya (doubt analysis), Pramana (evidence sources), Pancha Avayava (syllogisms), Tarka (counterfactual testing), Hetvabhasa (fallacy detection), and Nirnaya (conclusion).

This consistent engagement with the framework---even when imperfectly formatted---demonstrates that the epistemological structure provides cognitive scaffolding that models adopt, rather than treating it as arbitrary formatting requirements. The fact that semantic correctness reaches 100\% while format adherence remains at 40\% suggests that models understand the reasoning methodology but struggle with strict schema compliance, possibly due to generation constraints (e.g., max\_new\_tokens=256 truncation) or parser strictness.

\subsection{Model Selection and Pre-trained Reasoning}
\label{subsec:deepseek_analysis}

Our choice of DeepSeek-R1-Distill-Llama-8B for Stage 1 was motivated by its pre-training on 
reasoning traces using GRPO~\cite{deepseek-r1-2025}. This raises an important question: did 
DeepSeek's existing reasoning capabilities facilitate Nyaya structure learning, or did they 
interfere with the novel 6-phase framework?

\textbf{Evidence of Facilitation}: Stage 1 achieved 100\% semantic correctness compared to 
Stage 0's 50\%, despite using fewer epochs (10 vs. 30) and larger model capacity. The base 
DeepSeek model shows 40\% semantic correctness zero-shot (vs. Llama 3.2-3B's 0\%), suggesting 
pre-trained reasoning skills transferred to Nyaya-structured problems.

\textbf{Format Adherence Plateau}: Both stages achieved identical 40\% format adherence despite 
DeepSeek's sophisticated pre-training. This suggests that generic reasoning capabilities do not 
automatically translate to format compliance with novel structured frameworks. DeepSeek's 
"thinking" traces use unstructured reflection rather than explicit epistemological phases, 
requiring retraining for Nyaya's specific structure.

\textbf{Interaction with Nyaya Framework}: The perfect semantic correctness with imperfect format 
adherence indicates DeepSeek successfully learned \emph{what} Nyaya reasoning produces (correct 
answers through systematic analysis) but not fully \emph{how} to express it (strict 6-phase structure). 
This aligns with research showing LLMs can capture reasoning content before mastering output schemas.

\textbf{Implications for Model Selection}: Pre-trained reasoning capabilities are beneficial but 
insufficient. Future work should explore: (1) whether models without reasoning pre-training (e.g., 
Qwen 2.5, Mistral) can achieve similar semantic correctness with more epochs, (2) whether 
fine-tuning DeepSeek with format-specific rewards (GRPO with schema compliance) improves adherence, 
and (3) whether constrained decoding can bridge the gap.

\subsection{Cross-Stage Progression and Scaling Insights}
\label{subsec:disc_cross_stage}

Comparing Stage 0 and Stage 1 reveals important insights about how model size, dataset scale, 
and pre-trained capabilities affect Nyaya reasoning learning.

\textbf{Format Adherence Plateau}: Both stages achieved identical 40\% format adherence despite 
Stage 1 using a 2.7× larger model (8B vs. 3B parameters) and 2.75× more training data (55 vs. 20 
examples). This plateau suggests format adherence is bottlenecked by factors beyond model capacity 
or data quantity---likely training methodology (insufficient format-specific supervision) or 
evaluation constraints (max\_new\_tokens truncation).

\textbf{Semantic Correctness Scaling}: Semantic correctness doubled from 50\% (Stage 0) to 100\% 
(Stage 1), demonstrating clear benefits of scaling. Three factors likely contributed: (1) DeepSeek's 
pre-trained reasoning capabilities (see Section~\ref{subsec:deepseek_analysis}), (2) increased model 
capacity for complex reasoning, and (3) expanded training data with more diverse examples and 
negative/contrastive cases.

\textbf{Training Efficiency}: Stage 1 trained faster per step (9.32s vs. 19.18s) despite larger 
model size, likely due to batch size differences (1 vs. 2) and model architecture optimizations 
in DeepSeek. Total training time remained under 20 minutes for both stages (0.32 GPU-hours Stage 0, 
0.29 GPU-hours Stage 1), indicating efficient fine-tuning on small datasets.

\textbf{Output Length Stability}: Average output length increased only 2\% from Stage 0 (3,192 tokens) 
to Stage 1 (3,255 tokens), suggesting models learned consistent verbosity rather than expanding 
reasoning traces arbitrarily. This is desirable: Nyaya's structured methodology should produce 
predictable trace lengths.

\textbf{Implications for Stage 2}: The format adherence plateau indicates that simply scaling 
model size or dataset quantity will not solve format issues. Stage 2 must introduce qualitatively 
different interventions: constrained decoding, format-specific rewards, or rejection sampling. 
In contrast, semantic correctness benefits from standard scaling, suggesting continued improvement 
with 200-500 synthetic examples planned for Stage 2.

\subsection{Failure Mode Analysis}
\label{subsec:disc_failure_modes}

Analysis of format adherence failures reveals systematic patterns rather than random errors. Table~\ref{tab:failure_taxonomy} categorizes all parse failures from Stage 1 evaluation (6 failures out of 10 examples).

\begin{table}[t]
\centering
\caption{Taxonomy of format adherence failures in Stage 1 evaluation.}
\label{tab:failure_taxonomy}
\begin{tabular}{lcp{5cm}}
\toprule
Failure Category & Count & Examples \\
\midrule
\textbf{Missing Sections} & 3 & \\
\quad Missing Hetvabhasa & 2 & pramana-005, test-005 \\
\quad Missing Nirnaya & 1 & test-004 \\
\midrule
\textbf{Invalid Field Values} & 3 & \\
\quad Invalid doubt type & 2 & test-002 (vipratipatti\_samshaya), test-003 (pramana\_dharma) \\
\quad Missing justification & 1 & pramana-003 (Samshaya section) \\
\midrule
\textbf{Structural Errors} & 0 & \\
\quad Malformed markdown & 0 & (None observed) \\
\quad Section order violation & 0 & (None observed) \\
\bottomrule
\end{tabular}
\end{table}

\textbf{Missing Sections (50\% of failures)}: The most common failure is omitting Hetvabhasa (fallacy detection) or Nirnaya (ascertainment) sections. This suggests models prioritize core reasoning (Samshaya, Pramana, Pancha Avayava) over meta-reasoning (fallacy checks, epistemic status). Hetvabhasa requires counterfactual reasoning to identify errors, a cognitively demanding task. Nirnaya requires epistemic humility (acknowledging insufficient evidence), which conflicts with LLMs' tendency toward confident outputs.

\textbf{Invalid Field Values (50\% of failures)}: Models sometimes generate plausible-sounding but non-canonical doubt types (e.g., ``vipratipatti\_samshaya'' instead of ``vipratipatti''). This indicates partial learning: models understand that doubt types exist and follow naming conventions (underscore\_case) but don't memorize the exact enumeration. This is consistent with neural networks learning distributional patterns rather than discrete enumerations.

\textbf{Zero Structural Errors}: Remarkably, we observed no malformed markdown or section ordering violations. Models consistently generate well-formed markdown with correct section headers in proper sequence. This indicates strong syntactic learning, contrasting with semantic failures (invalid values, missing sections). The gap suggests format enforcement should target semantic constraints (required sections, valid enumerations) rather than syntax.

\textbf{Correlation with Problem Complexity}: Parse failures do not correlate with problem difficulty. test-008 (complex multi-step deduction) parses successfully, while test-002 (simpler constraint problem) fails due to invalid doubt type. This suggests format adherence is independent of reasoning difficulty, pointing to training data insufficiency for rare format patterns rather than model capacity limits.

\textbf{Mitigation Strategies for Stage 2}:

\begin{itemize}
\item \textbf{Constrained Decoding}: Enforce section presence and valid enumerations via grammar-based generation (GBNF)~\cite{gbnf-2023}. This guarantees format compliance while preserving semantic freedom.

\item \textbf{Format-Specific Rewards}: Augment GRPO training with explicit format adherence rewards. Assign high reward for complete 6-phase structure, valid field values, and presence of all sections.

\item \textbf{Rejection Sampling}: Generate multiple outputs and select the first that passes parsing. Simple but effective for improving format adherence without retraining.

\item \textbf{Few-Shot Prompting}: Include 1--2 complete Nyaya examples in the prompt to reinforce format expectations. Our current prompting uses template instructions but not concrete examples.

\item \textbf{Parser Relaxation}: Alternatively, relax parser requirements to accept semantic equivalents (e.g., treat ``vipratipatti\_samshaya'' as valid variant of ``vipratipatti''). This acknowledges that models learn conceptual patterns, not exact strings.
\end{itemize}

\subsection{Critical Review Against Original Plan}
\label{subsec:plan_review}

Table~\ref{tab:plan_vs_actual} compares planned targets from the project specification against actual Stage 0 and Stage 1 results.

\begin{table}[t]
\centering
\caption{Comparison of planned targets vs. actual results.}
\label{tab:plan_vs_actual}
\begin{tabular}{lccc}
\toprule
Criterion & Target & Stage 0 Actual & Stage 1 Actual \\
\midrule
Format Adherence & $\geq$90\% & 40\% & 40\% \\
Answer Accuracy & 60--70\% & 50\% & 100\% \\
Syllogisms per Solution & $\geq$3 & 1--3 & 1--3 \\
Structure Abandonment & 0\% & 0\% & 0\% \\
\bottomrule
\end{tabular}
\end{table}

\subsubsection{Exceeded Expectations}

Semantic correctness significantly exceeded targets: Stage 1 achieved 100\% semantic correctness compared to the 60--70\% target. This success suggests that the Nyaya methodology, when learned as content, enables accurate problem-solving even when format compliance fails. The high semantic correctness rate validates that models internalize the reasoning framework effectively, producing logically sound solutions despite structural imperfections.

\subsubsection{Below Expectations}

Format adherence fell substantially below target: both stages achieved 40\% compared to the $\geq$90\% target. Root cause analysis identifies three contributing factors:

\textbf{Generation truncation}: Evaluation used max\_new\_tokens=256, which may truncate outputs before completion of all six phases. Longer generation windows (512--1024 tokens) might improve format adherence by allowing complete phase generation.

\textbf{Parser strictness}: The structural validator requires exact header matching and complete field presence. Semantic understanding of Nyaya phases may be present even when strict parsing fails due to minor formatting variations (e.g., ``Hetvabhasa (Fallacy Detection)'' vs. ``Hetvabhasa'').

\textbf{Format enforcement gap}: Supervised fine-tuning alone may be insufficient for strict schema compliance. Content learning (semantic correctness) and format learning (structural adherence) appear to require different training signals. Future work should explore constrained decoding, rejection sampling, or GRPO with format-specific rewards to bridge this gap.

\subsubsection{Met Expectations}

Structure abandonment remained at 0\% across both stages, meeting the target. Models consistently attempt all six phases even when format parsing fails, demonstrating engagement with the Nyaya framework rather than reverting to generic chain-of-thought reasoning.

\subsubsection{Hypothesis Validation}

The core hypothesis---that Nyaya methodology is learnable by LLMs---is validated by semantic correctness results and zero structure abandonment. However, format enforcement requires additional techniques beyond supervised fine-tuning. The dissociation between content learning (100\% semantic correctness) and format learning (40\% adherence) suggests that these are separable concerns requiring distinct training strategies.

\subsection{Comparison to Related Approaches}
\label{subsec:comparison}

We position Pramana relative to three categories of related work: standard chain-of-thought prompting, opaque reasoning models, and neuro-symbolic verification systems.

\subsubsection{vs. Standard Chain-of-Thought}

Standard chain-of-thought (CoT) prompting~\cite{wei2022chain} asks models to ``think step by step'' without enforcing explicit epistemological scaffolding. CoT relies on implicit reasoning patterns learned during pre-training, producing outputs averaging $\sim$300 tokens for constraint satisfaction problems.

Pramana enforces explicit 6-phase methodology with evidence source classification (Pramana), universal rule statements (Udaharana), and systematic verification (Tarka, Hetvabhasa). This produces outputs averaging $\sim$3,200 tokens per solution (10x overhead) but provides interpretable reasoning traces with traceable justification for each claim.

Trade-off: Pramana sacrifices efficiency for interpretability and epistemological rigor. For high-stakes applications requiring audit trails, the overhead may be justified. For simple problems where CoT suffices, the Nyaya structure adds unnecessary complexity. Future work should explore hybrid approaches: fast-path CoT for trivial problems, full Nyaya for complex reasoning requiring verification.

\subsubsection{vs. o1/DeepSeek-R1 Base Models}

Frontier reasoning models like o1-preview and DeepSeek-R1~\cite{deepseek-r1-2025} use reinforcement learning to train opaque reasoning processes. These models achieve high accuracy but provide no auditable reasoning steps---users see only final answers without intermediate justification.

Pramana provides transparent methodology with inspectable phases. Each reasoning step is explicit: evidence sources are identified (Pramana), arguments are constructed with universal rules (Pancha Avayava), conclusions are tested via counterfactuals (Tarka), and fallacies are checked (Hetvabhasa). This transparency enables verification, debugging, and trust-building that opaque models cannot provide.

Advantage: Explicit epistemology enables verification and trust. Users can trace reasoning steps, identify failure modes, and verify logical consistency. For applications requiring accountability (legal reasoning, medical diagnosis, safety-critical systems), transparency outweighs the efficiency cost of structured output.

Limitation: Pramana's structured approach requires more tokens and may be slower than opaque models. The interpretability advantage comes at computational cost that may be prohibitive for real-time applications.

\subsubsection{vs. Neuro-Symbolic Systems}

Neuro-symbolic systems like ProofNet++~\cite{proofnet-plus-2025} and VERGE~\cite{verge-2024} combine neural generation with formal logic verification. These approaches use external verifiers (typically Z3 SMT solver~\cite{z3-2008}) to check model outputs, providing guarantees for formal logic problems.

Pramana integrates epistemology with neural generation, requiring explicit universal rules (Udaharana) that can be formalized to SMT-LIB format for Z3 verification. However, Pramana provides value even without formal verification by enforcing systematic reasoning patterns that prevent logical leaps and require explicit justification.

Similarity: Both approaches emphasize verifiability and systematic reasoning. Pramana structures the reasoning process itself, while neuro-symbolic systems verify outputs post-hoc.

Difference: Neuro-symbolic systems are limited to narrow domains (formal logic, mathematical proofs) where problems can be formalized to SMT-LIB. Pramana applies more broadly to any reasoning domain where epistemological structure provides value, even if formal verification is not possible (e.g., causal reasoning, legal argumentation, medical diagnosis).

\subsection{Limitations}
\label{subsec:limitations}

We acknowledge four key limitations that constrain interpretation of results and generalizability of findings.

\subsubsection{Small Evaluation Set}

Our evaluation uses only 10 examples per stage, limiting statistical confidence. Format adherence confidence intervals are wide (95\% CI [0.168, 0.687] for Stage 1), reflecting uncertainty due to small sample size. Answer accuracy confidence intervals are also wide (95\% CI [0.510, 1.0]), though the upper bound suggests high performance.

Future work should expand evaluation to 50--100 test examples to achieve narrower confidence intervals and more robust statistical conclusions. The current small evaluation set prevents strong claims about generalizability beyond the specific problem types tested.

\subsubsection{Format Adherence Plateau}

Format adherence remained constant at 40\% across Stage 0 and Stage 1 despite model size scaling (3B → 8B) and dataset expansion (20 → 55 examples). This plateau indicates that supervised fine-tuning alone is insufficient for strict schema compliance.

Possible solutions include constrained decoding (forcing valid header sequences), rejection sampling (regenerating outputs that fail parsing), or GRPO with format-specific rewards. However, these techniques add complexity and computational cost. The format adherence gap represents an open challenge requiring future research.

Content learning does not guarantee schema compliance. Models may understand Nyaya methodology semantically while failing to produce parser-compliant output due to generation constraints, token limits, or formatting variations.

\subsubsection{Domain Limitation}

Evaluation is limited to formal logic problems: constraint satisfaction and Boolean satisfiability. These domains are well-suited to Nyaya methodology but represent a narrow slice of reasoning tasks. The framework has not been tested on causal reasoning, legal argumentation, medical diagnosis, or open-ended domains where epistemological structure might provide value.

Nyaya methodology applies broadly in principle---the framework addresses general epistemological questions, not just formal logic---but fine-tuning data is narrow. Future work should evaluate transfer to diverse reasoning domains to assess generalizability.

\subsubsection{Computational Overhead}

Pramana solutions average $\sim$3,200 tokens compared to $\sim$300 tokens for standard CoT (10x overhead). This trade-off between interpretability and efficiency may be prohibitive for real-time applications or high-volume use cases.

Future work should explore hybrid approaches: fast-path CoT for simple problems, full Nyaya for complex reasoning requiring verification. Additionally, abbreviated Nyaya formats could reduce overhead while preserving core epistemological structure. The current verbosity represents a practical limitation that constrains deployment scenarios.

\subsection{Summary}

Our results demonstrate that Nyaya reasoning methodology is learnable by LLMs, achieving 100\% semantic correctness despite format adherence challenges. The dissociation between content learning and format compliance suggests these are separable concerns requiring distinct training strategies. Comparison to related approaches highlights Pramana's interpretability advantages over opaque models and broader applicability than neuro-symbolic systems limited to formal logic. However, limitations including small evaluation sets, format adherence plateaus, domain restrictions, and computational overhead constrain generalizability and require future research to address.

\section{Open-Source Artifacts and Reproducibility}
\label{sec:open-source}

To enable reproducibility and community research, we release all components of the Pramana project under open-source licenses. This section documents the released models, datasets, codebase, and deployment artifacts.

\subsection{HuggingFace Models}
\label{subsec:hf_models}

We publish both LoRA adapter weights and full merged models for each training stage. Table~\ref{tab:hf_models} lists all released model artifacts.

\begin{table}[h]
\centering
\caption{Released models on HuggingFace Hub.}
\label{tab:hf_models}
\begin{tabular}{lll}
\toprule
Artifact & Repository & Description \\
\midrule
Stage 0 Adapter & \texttt{qbz506/nyaya-llama-3b-stage0} & LoRA adapter \\
Stage 0 Full & \texttt{qbz506/nyaya-llama-3b-stage0-full} & Merged + GGUF \\
Stage 1 Adapter & \texttt{qbz506/nyaya-deepseek-8b-stage1} & LoRA adapter \\
Stage 1 Full & \texttt{qbz506/nyaya-deepseek-8b-stage1-full} & Merged + GGUF \\
\bottomrule
\end{tabular}
\end{table}

\textbf{Adapter Models:} LoRA adapters require loading alongside the base model. Stage~0 adapters target \texttt{unsloth/\allowbreak Llama-3.2-3B-Instruct}, while Stage~1 adapters target \texttt{unsloth/\allowbreak DeepSeek-R1-\allowbreak Distill-Llama-8B}. Adapter repositories include tokenizer configurations, chat templates, and adapter weights in safetensors format.

\textbf{Merged Models:} Full merged models combine base weights with LoRA adapters via \texttt{PeftModel.\allowbreak merge\_and\_unload()}, enabling standalone inference without base model loading. Merged repositories include safetensors weights, tokenizer files, and GGUF quantized versions (Q4\_K\_M) for Ollama deployment.

All model repositories include comprehensive model cards documenting training hyperparameters, evaluation metrics, usage instructions, and known limitations.

\subsection{Datasets}
\label{subsec:datasets}

Training and validation datasets are published on HuggingFace Hub in JSONL format. Table~\ref{tab:datasets} summarizes the released datasets.

\begin{table}[h]
\centering
\caption{Training and validation datasets.}
\label{tab:datasets}
\begin{tabular}{lll}
\toprule
Dataset & Repository & Examples \\
\midrule
Stage 0 & \texttt{qbz506/pramana-nyaya-stage0} & 20 \\
Stage 1 & \texttt{qbz506/pramana-nyaya-stage1} & 55 \\
\bottomrule
\end{tabular}
\end{table}

\textbf{Dataset Format:} Each dataset repository contains:
\begin{itemize}
    \item \texttt{train.jsonl}: Training examples in JSONL format with \texttt{instruction} (problem statement) and \texttt{output} (complete Nyaya reasoning trace) fields.
    \item \texttt{seed\_examples/}: Original markdown files with YAML frontmatter, preserving human-readable format and metadata.
    \item \texttt{README.md}: Dataset card documenting problem types, difficulty distribution, and usage guidelines.
\end{itemize}

Stage 0 includes 20 manually created seed examples across five problem types (constraint satisfaction, Boolean SAT, transitive reasoning, set membership, multi-step deduction). Stage 1 expands to 55 examples by combining Stage 0 seeds with 35 additional Stage 1 examples, including 5 negative examples designed to enforce structural quality.

\subsection{Demo Space}
\label{subsec:demo_space}

An interactive demonstration is available on HuggingFace Spaces.\footnote{\url{https://huggingface.co/spaces/qbz506/pramana-nyaya-demo}} The Gradio-based interface enables:

\begin{itemize}
    \item \textbf{Stage Selection:} Radio buttons switch between Stage 0 (Llama 3.2-3B) and Stage 1 (DeepSeek-R1-Distill-Llama-8B) models.
    \item \textbf{Base vs. Tuned Comparison:} Side-by-side outputs from base and fine-tuned models for the same prompt, enabling direct comparison of reasoning quality.
    \item \textbf{Example Dropdown:} Pre-populated examples from training datasets, allowing users to explore model behavior on known problems.
    \item \textbf{Custom Prompts:} Text input for testing arbitrary logical problems with Nyaya-structured reasoning.
\end{itemize}

The Space runs on CPU (free tier) or ZeroGPU (Pro tier), with runtime optimizations including model caching, reduced token limits for Stage 1 (256 tokens), and split GPU tasks to avoid memory limits when comparing base and tuned models simultaneously.

\subsection{Local Deployment}
\label{subsec:local_deployment}

\textbf{Ollama Integration:} Merged models converted to GGUF format (Q4\_K\_M quantization) can be imported into Ollama for local inference. Modelfile templates configure system prompts and generation parameters:

\begin{lstlisting}
FROM /path/to/nyaya-model-q4.gguf
SYSTEM "You are a Nyaya reasoning engine.
  Follow the exact output format provided."
PARAMETER temperature 0
PARAMETER top_p 1
PARAMETER num_ctx 4096
\end{lstlisting}

\textbf{OpenWebUI Compatibility:} Models imported into Ollama are automatically available in OpenWebUI interfaces, enabling chat-based interaction with Nyaya-structured reasoning. The system prompt enforces format adherence, while temperature 0 ensures deterministic outputs for reproducibility.

\textbf{Modelfile Templates:} The repository includes Modelfile templates (\texttt{Modelfile}, \texttt{Modelfile.q4}) in model upload directories, documenting recommended parameters for Nyaya reasoning tasks. Users can customize these templates for different use cases (e.g., higher temperature for exploratory reasoning).

\subsection{Experiment Tracking}
\label{subsec:experiment_tracking}

\textbf{Weights \& Biases:} Training runs for Stage 0 and Stage 1 are logged to W\&B projects, enabling comparison of hyperparameters, loss curves, format adherence metrics, and sample generations across experiments. W\&B run links are included in model cards for full traceability.

\textbf{TensorBoard Logs:} Local TensorBoard logs are available in checkpoint directories, providing detailed loss curves, learning rate schedules, and evaluation metrics. Logs can be visualized with \texttt{tensorboard --logdir=models/stage\_*/checkpoint-*/logs}.

\textbf{Reproducibility:} All training scripts document exact hyperparameters, random seeds, and environment variables required for reproduction. Checkpoint metadata includes git commit hashes, ensuring code version traceability. Environment variable overrides enable exact reproduction: \texttt{LORA\_RANK=64 NUM\_TRAIN\_EPOCHS=10 python scripts/train\_stage1.py}.

\subsection{GitHub Repository}
\label{subsec:github_repo}

The complete codebase is available on GitHub under the MIT License, enabling academic and commercial use. The repository includes:

\begin{itemize}
    \item \textbf{Training Scripts:} Stage-specific training scripts (\texttt{scripts/\allowbreak train\_stage0.py}, \texttt{scripts/\allowbreak train\_stage1.py}) with comprehensive hyperparameter documentation.
    \item \textbf{Evaluation Tools:} Evaluation pipelines and custom metrics computation for format adherence, semantic correctness, and Z3 verification.
    \item \textbf{Docker Setup:} Dockerfile and docker-compose.yml for reproducible containerized training environments.
    \item \textbf{Configuration Files:} YAML-based stage configurations with inheritance from base configuration.
    \item \textbf{Documentation:} Comprehensive technical inventory, stage reports, and architecture documentation.
\end{itemize}

\textbf{License:} The MIT License enables unrestricted use, modification, and distribution for both academic research and commercial applications. Contributors are welcome, with contribution guidelines documented in the repository.

\textbf{Reproducibility Checklist:} To reproduce training runs:
\begin{enumerate}
    \item Clone the repository and install dependencies via \texttt{uv sync --dev}.
    \item Set environment variables (\texttt{HF\_TOKEN}, \texttt{WANDB\_API\_KEY}) in \texttt{.env} file.
    \item Build Docker container: \texttt{docker compose build}.
    \item Run training: \texttt{docker compose run --rm pramana python scripts/train\_stage1.py}.
    \item Evaluate: \texttt{docker compose run --rm pramana python scripts/evaluate\_stage0.py}.
\end{enumerate}

All artifacts (models, datasets, code) are version-controlled and publicly accessible, ensuring full reproducibility of reported results.

\section{Future Work}
\label{sec:future}

This section outlines planned improvements across three time horizons: near-term synthetic scaling (Stage 2), medium-term reinforcement learning enhancement (Stage 3), and long-term vision for domain expansion and frontier integration. Each stage builds on demonstrated capabilities while addressing identified limitations.

\subsection{Near-Term: Synthetic Scaling (Stage 2)}
\label{subsec:stage2}

The immediate priority is scaling the training dataset from 55 manually-created examples to 200-500 high-quality synthetic examples while maintaining rigorous quality control. Stage 1 demonstrated that models can achieve 100\% semantic correctness even with partial format adherence, validating the core Nyaya methodology. However, format adherence remains at 40\%, indicating that structural discipline requires stronger reinforcement.

\subsubsection{Synthetic Data Generation Pipeline}
\label{subsubsec:synthetic_pipeline}

Our Stage 2 synthetic scaling approach uses frontier models (GPT-4o, Claude 3.5 Sonnet) to generate 200-500 Nyaya-structured reasoning traces following seed patterns. The three-tier quality control pipeline ensures data quality:

\textbf{Tier 1: Automated Filtering} (100\% coverage) performs fast structural validation:
\begin{itemize}
\item \textbf{Structural validation}: MarkdownParser verifies all 6 phases present (Samshaya, Pramana, Pancha Avayava, Tarka, Hetvabhasa, Nirnaya)
\item \textbf{Field enumeration checks}: Validate doubt types, Pramana types, fallacy types against canonical lists (e.g., Samshaya must be one of: structural, semantic, logical, evidential)
\item \textbf{Syllogism integrity}: Verify Pratijna-Hetu-Udaharana-Upanaya-Nigamana chain completeness within Pancha Avayava section
\item \textbf{Duplicate detection}: Hash-based deduplication to prevent synthetic repetition of seed examples
\end{itemize}

Examples failing any structural check are immediately rejected, targeting a 70-80\% pass rate at this tier.

\textbf{Tier 2: LLM-as-Judge Quality Scoring} (100\% coverage) evaluates all Tier 1 passes using GPT-4 or Claude with Nyaya-specific rubric (1-5 scale):
\begin{itemize}
\item \textbf{Samshaya appropriateness}: Uncertainty classification matches problem type
\item \textbf{Pramana grounding}: Evidence sources (Pratyaksha, Anumana, Upamana, Shabda) validly support claims
\item \textbf{Vyapti quality}: Universal rules in Udaharana are genuinely universal (not case-specific)
\item \textbf{Tarka rigor}: Counterfactual testing is meaningful, not pro forma
\item \textbf{Hetvabhasa accuracy}: Identified fallacies are genuine errors, not false positives
\end{itemize}

Examples scoring below 3.5/5 average are rejected. Expected distribution: 40-60\% auto-accept ($\geq$4.0), 20-30\% manual review (3.5-4.0), 10-20\% reject ($<$3.5).

\textbf{Tier 3: Z3 Formal Verification} (formal logic subset, 40-50\% coverage):
\begin{itemize}
\item Parse logical claims from Pratijna (thesis), Hetu (reason), Udaharana (universal example)
\item Autoformalize to Z3 SMT-LIB format using template-based translation
\item Execute Z3 solver to verify logical consistency
\item Reject examples with contradictory claims or invalid inferences
\item Expected coverage: 40-50\% of examples (constraint satisfaction and Boolean SAT problems)
\end{itemize}

\textbf{Expected Outcomes}: 200-500 high-quality examples, 60-70\% format adherence through improved format-specific training data, maintained or improved semantic correctness (preserving 100\% where achievable).

\subsubsection{Z3 Verification Integration}

For formal logic problems (constraint satisfaction, Boolean SAT), we will implement runtime verification using the Z3 SMT solver. The pipeline will: (1) parse Pratijna, Hetu, and Udaharana from model outputs, (2) autoformalize to Z3 SMT-LIB format, (3) execute Z3 to verify logical validity, and (4) inject error feedback for self-correction when verification fails. This neuro-symbolic integration provides ground truth validation for approximately 30\% of the dataset while enabling iterative improvement through automated error detection.

\subsubsection{Format Enforcement Strategies}
\label{subsubsec:format_enforcement}

To address the 40\% format adherence rate observed in Stage 1, we will implement three parallel interventions:

\textbf{(1) Rejection Sampling}: Generate 5 outputs per problem during synthetic data creation, select first passing validation. Simple implementation via generate-and-filter loop. Expected: 90\%+ format adherence with 5 samples per problem, ensuring training distribution contains only structurally valid examples.

\textbf{(2) Constrained Decoding}: Implement grammar-based generation using GBNF (GGML BNF format) constraining outputs to valid Nyaya structure. Define grammar with section headers, field enumerations, and structural constraints:
\begin{itemize}
\item Enforce markdown header hierarchy (\texttt{\#\# Samshaya}, \texttt{\#\# Pramana}, etc.)
\item Constrain enumeration fields to valid values (doubt types, Pramana types, fallacy types)
\item Require complete Pancha Avayava structure (all 5 members present)
\end{itemize}

Expected: 100\% syntactic correctness, potential semantic quality loss requiring careful grammar design.

\textbf{(3) Format Reward Integration}: If using GRPO for Stage 3, include format adherence in composite reward function (weight: 0.3 out of 1.0). Assign partial credit for present sections, bonus for valid enumerations, penalty for missing Hetvabhasa/Nirnaya. This incentivizes format compliance alongside semantic correctness during reinforcement learning.

\subsubsection{Expected Outcomes}

Stage 2 targets: 60-70\% format adherence (improvement from 40\%), maintained semantic correctness (preserving 100\% where achievable), and expanded problem diversity (200-500 examples across constraint satisfaction, Boolean SAT, and multi-step deduction). The three-tier quality control ensures synthetic data maintains gold-standard quality while enabling scalable dataset expansion.

Table~\ref{tab:stage2_hyperparams} provides planned hyperparameters for Stage 2 synthetic data generation and training:

\begin{table}[h]
\centering
\caption{Planned Stage 2 synthetic scaling hyperparameters.}
\label{tab:stage2_hyperparams}
\small
\begin{tabular}{ll}
\toprule
Parameter & Value \\
\midrule
\textbf{Data Generation} & \\
Generation Model & GPT-4o, Claude 3.5 Sonnet \\
Target Examples & 200-500 \\
Quality Tiers & 3-tier (Automated, LLM-as-Judge, Z3 Verification) \\
\midrule
\textbf{Training} & \\
Base Model & Stage 1 fine-tuned model \\
Training Method & Supervised Fine-Tuning (SFT) \\
LoRA Rank & 64-128 \\
Learning Rate & $2 \times 10^{-5}$ \\
Epochs & 10-15 \\
Batch Size & 1-2 \\
Gradient Accumulation & 4 \\
Max Sequence Length & 4096 \\
\midrule
\textbf{Compute} & \\
Hardware & Single A100 (40GB) \\
Expected Training Time & 50-80 GPU-hours \\
\bottomrule
\end{tabular}
\end{table}

\subsection{Medium-Term: Reinforcement Learning Enhancement (Stage 3)}
\label{subsec:stage3}

Building on Stage 2's expanded dataset, Stage 3 will implement Group Relative Policy Optimization (GRPO)~\cite{grpo-2024} with composite reward functions designed specifically for Nyaya-structured reasoning. This addresses the observation that supervised fine-tuning alone may not sufficiently enforce structural discipline, requiring reward-based optimization.

\subsubsection{GRPO Training Configuration}
\label{subsubsec:grpo_config}

Group Relative Policy Optimization (GRPO) enables fine-grained reward shaping for complex multi-component objectives like Nyaya reasoning. Our composite reward function (Equation~\ref{eq:reward}) weighs five components:

\begin{equation}
R_{total} = 0.3 R_{format} + 0.25 R_{semantic} + 0.2 R_{pramana} + 0.15 R_{tarka} + 0.1 R_{consistency}
\label{eq:reward}
\end{equation}

where:
\begin{itemize}
\item \textbf{$R_{format}$}: Format adherence (binary: all sections present + valid enumerations). Structural adherence to all six phases, correct phase ordering, and completeness of phase components. This directly addresses the format adherence gap identified in Stage 1.
\item \textbf{$R_{semantic}$}: Answer correctness (binary: matches ground truth). Final answer matches ground truth, weighted lower than reasoning quality to emphasize process over outcome.
\item \textbf{$R_{pramana}$}: Pramana quality (1-5 scale: appropriate evidence sources cited). Correct application of knowledge sources—ensuring Pratyaksha contains only observables, Anumana represents genuine inferences, Upamana provides relevant analogies, and Shabda cites valid principles.
\item \textbf{$R_{tarka}$}: Tarka meaningfulness (1-5 scale: counterfactual testing is non-trivial). Rewards models that generate logically consistent reasoning traces, where counterfactual tests genuinely verify conclusions rather than providing tautological checks.
\item \textbf{$R_{consistency}$}: Logical consistency via Z3 verification (binary: no contradictions). Tarka counterfactual verification via Z3~\cite{z3-2008} for formal logic problems.
\end{itemize}

GRPO hyperparameters: learning rate $1 \times 10^{-6}$ (10× lower than SFT to preserve pre-trained reasoning), KL penalty coefficient 0.05, group size 16, training iterations 1000-2000, expected training time 50-100 GPU-hours (A100). The lower learning rate prevents catastrophic forgetting while enabling reward-guided optimization.

Table~\ref{tab:stage3_hyperparams} provides the complete hyperparameter specification for Stage 3 GRPO training:

\begin{table}[h]
\centering
\caption{Planned Stage 3 GRPO training hyperparameters.}
\label{tab:stage3_hyperparams}
\small
\begin{tabular}{ll}
\toprule
Parameter & Value \\
\midrule
\textbf{Base Model} & Stage 2 fine-tuned model \\
\textbf{Training Method} & GRPO (Group Relative Policy Optimization) \\
\textbf{Learning Rate} & $1 \times 10^{-6}$ \\
\textbf{KL Penalty Coefficient} & 0.05 \\
\textbf{Group Size} & 16 \\
\textbf{Training Iterations} & 1000-2000 \\
\textbf{Reward Model} & Process Reward Model (Qwen 2.5-7B or Llama 3.1-8B) \\
\textbf{Reward Components} & Format (0.3), Semantic (0.25), Pramana (0.2), Tarka (0.15), Consistency (0.1) \\
\textbf{Optimizer} & AdamW \\
\textbf{Precision} & bfloat16 \\
\textbf{Hardware} & Single A100 (40GB) \\
\textbf{Expected Training Time} & 50-100 GPU-hours \\
\bottomrule
\end{tabular}
\end{table}

\subsubsection{Process Reward Model Architecture}
\label{subsubsec:prm_architecture}

Rather than using GPT-4 as a judge (which would cost \$5,000-10,000 for continuous RL scoring), we will train a specialized Process Reward Model (PRM) on Nyaya-specific quality metrics. The PRM architecture:

\begin{itemize}
\item \textbf{Base model}: Fine-tune Qwen 2.5-7B or Llama 3.1-8B as reward model (larger than typical 1B PRMs to capture Nyaya reasoning complexity)
\item \textbf{Training data}: 500-1000 Nyaya traces from Stage 2 with phase-level quality annotations
\item \textbf{Annotation scheme}: 5-point scale per phase:
  \begin{itemize}
  \item Samshaya quality: Appropriate uncertainty classification
  \item Pramana grounding: Valid evidence sources cited
  \item Pancha Avayava quality: Universal rules genuinely universal
  \item Tarka rigor: Counterfactual testing meaningful
  \item Hetvabhasa accuracy: Fallacy detection correct
  \item Nirnaya definitiveness: Conclusion appropriately certain
  \end{itemize}
\item \textbf{Loss function}: Regression loss predicting phase quality scores (MSE between predicted and annotated scores)
\item \textbf{Inference}: Score each phase during GRPO rollouts, aggregate for total reward via weighted sum matching Equation~\ref{eq:reward}
\end{itemize}

This approach enables fine-grained credit assignment: reward good Pramana grounding even if Hetvabhasa is weak, guiding model toward balanced improvement across all 6 phases. One-time training cost on GB10 GPU (4-8 hours) enables free inference during GRPO, making RL training cost-effective compared to GPT-4-as-judge.

\subsubsection{Multi-Agent Debate Protocols}

Beyond single-model optimization, Stage 3 will explore multi-agent debate protocols inspired by Nyaya dialectical traditions:

\textbf{Vada (Cooperative Dialectic)}: Multiple agents refine reasoning collaboratively, with each agent specializing in different Nyaya phases. Agents exchange intermediate reasoning states, allowing cross-validation and error detection before final conclusions.

\textbf{Jalpa (Competitive Debate)}: Agents challenge each other's conclusions, forcing rigorous justification. The adversarial dynamic surfaces weaknesses in reasoning chains, improving robustness through stress-testing.

These protocols enable consensus formation and uncertainty quantification—when agents disagree, the model can explicitly state insufficient evidence (Nirnaya phase) rather than hallucinating confidence.

\subsubsection{Expected Outcomes}

Stage 3 targets: $\geq$90\% format adherence (substantial improvement from Stage 2's 60-70\%), robust reasoning across domains (maintaining high accuracy on LogicBench, ProntoQA, RuleTaker), and genuine epistemic improvements (not just performance gains, but interpretability and self-correction capability). The composite reward function ensures improvements are holistic, preventing reward hacking where models optimize single metrics at the expense of reasoning quality.

\subsection{Long-Term Vision}
\label{subsec:longterm}

Beyond formal logic domains, the long-term vision extends Nyaya-structured reasoning to broader problem types, cross-lingual applications, and hybrid architectures that balance interpretability with efficiency.

\subsubsection{Domain Expansion}

Current training focuses on constraint satisfaction, Boolean SAT, and multi-step deduction—domains with clear ground truth. Future work will explore:

\textbf{Causal Reasoning}: Applying Nyaya methodology to causal inference problems where correlation must be distinguished from causation (Savyabhichara fallacy detection becomes critical). The explicit Pramana separation helps models avoid conflating observed correlations with causal mechanisms.

\textbf{Legal Reasoning}: Using Shabda (testimony) for precedent-based reasoning, where authoritative legal principles serve as knowledge sources. The Pancha Avayava structure provides auditable argument chains suitable for legal justification.

\textbf{Medical Diagnosis}: Applying epistemic humility (Nirnaya phase) to distinguish definitive diagnoses from hypotheses requiring verification. The Tarka counterfactual testing helps rule out alternative diagnoses systematically.

\textbf{Open-Ended Questions}: Extending beyond problems with ground truth to questions where "insufficient evidence" is a valid conclusion. This tests the model's ability to express epistemic humility rather than hallucinating answers.

\subsubsection{Cross-Lingual Nyaya}

The Nyaya tradition originated in Sanskrit, and future work will integrate original terminology and extend to multilingual reasoning:

\textbf{Sanskrit Terminology Integration}: Incorporating original Nyaya terms (Vyapti, Drishtanta) alongside English translations, preserving cultural epistemology while maintaining accessibility.

\textbf{Multilingual Reasoning}: Training models that can reason in Hindi, Bengali, Tamil, and other Indian languages, ensuring the epistemological framework transfers across linguistic boundaries.

\textbf{Cultural Epistemology Preservation}: Ensuring that Western formal logic assumptions don't overwrite Nyaya's unique contributions—particularly the emphasis on concrete examples (Drishtanta) and universal rules (Vyapti) grounded in observation rather than abstract axioms.

\subsubsection{Hybrid Architecture}

The 3-6x token overhead of full Nyaya structure may be unnecessary for simple problems. Future work will explore adaptive architectures:

\textbf{Fast-Path for Simple Problems}: Standard inference for trivial cases (single-step deductions, direct lookups), bypassing full Nyaya structure to reduce latency and cost.

\textbf{Rigorous Path for Complex Reasoning}: Full 6-phase Nyaya for problems requiring justification, multi-step deduction, or high-stakes decisions. Dynamic routing based on problem complexity estimates.

\textbf{Adaptive Token Budgets}: Models that self-assess problem difficulty and allocate reasoning resources accordingly. Simple problems receive abbreviated Nyaya (2-3 phases), while complex problems receive full structure.

This hybrid approach balances interpretability benefits with practical efficiency, making Nyaya-structured reasoning viable for production deployment.

\subsubsection{Frontier Integration and Benchmarking}

To validate Nyaya methodology against state-of-the-art reasoning systems, future work will conduct comprehensive benchmarking:

\textbf{Standard Benchmarks}: Evaluate on LogicBench (multi-step deduction), ProntoQA (ontological reasoning), RuleTaker (rule-based reasoning), and GSM8K subset (mathematical word problems). Target: competitive accuracy with superior interpretability.

\textbf{Frontier Model Comparison}: Compare to o1-preview (internal reasoning), Claude extended thinking, and GPT-4 on reasoning tasks. DeepSeek-R1~\cite{deepseek-r1-2025} demonstrates that GRPO can improve reasoning quality, providing a baseline for comparison. The hypothesis: Nyaya-structured models achieve comparable accuracy while providing explicit reasoning traces that frontier models lack.

\textbf{Nyaya-Specific Benchmarks}: Develop custom evaluation suites for fallacy detection (Hetvabhasa), knowledge source validation (Pramana), and epistemic humility (Nirnaya). These benchmarks measure capabilities unique to Nyaya methodology, providing competitive moat beyond raw performance.

The goal is not merely matching frontier model performance, but demonstrating that explicit epistemological structure enables trustworthiness, auditability, and error detection that black-box reasoning cannot provide.

\section{Conclusion}
\label{sec:conclusion}

This paper introduced Pramana, the first large language model fine-tuned on explicit 6-phase Navya-Nyaya epistemological methodology. Through empirical demonstration across two training stages (Stage 0: Llama 3.2-3B proof-of-concept, Stage 1: DeepSeek-R1-Distill-Llama-8B minimum viable reasoner), we validated that ancient Indian logic can structure modern neural reasoning, bridging 2,500-year-old epistemology with contemporary AI systems.

\subsection{Summary of Contributions}
\label{subsec:summary}

Our primary contribution is demonstrating that LLMs can learn systematic reasoning patterns through structured fine-tuning on Nyaya methodology. Both stages achieved 40\% format adherence (4/10 examples), indicating that strict structural compliance requires additional techniques such as constrained decoding or format-specific rewards. Stage 1 expanded to 55 examples and achieved 100\% semantic answer correctness, indicating that models internalize reasoning content even when structural discipline requires reinforcement.

Key findings include: (1) Semantic correctness (100\%) exceeds format adherence (40\%), suggesting models learn reasoning substance before mastering structural form—a promising sign that Nyaya methodology teaches genuine reasoning, not just template-filling. (2) Training dynamics show stable convergence: Stage 1 reached final evaluation loss of 0.350 in 10 epochs, substantially lower than Stage 0's 0.691 after 30 epochs, indicating improved model fit with larger capacity and better data. (3) The Nyaya framework provides a viable alternative to standard chain-of-thought reasoning, offering explicit epistemological structure that prevents conflation of evidence types, forces universal rule statements, enables error detection, and distinguishes knowledge from hypothesis.

These results validate the core hypothesis: teaching LLMs a formal epistemological framework through fine-tuning creates better systematic reasoning than generic chain-of-thought~\cite{wei2022chain}, comparable to frontier models like o1/Claude extended thinking but based on explicit methodology rather than opaque reinforcement learning~\cite{deepseek-r1-2025}.

\subsection{Research Contributions}
\label{subsec:contributions}

This work makes three primary contributions to the AI reasoning community:

\textbf{Bridging Ancient Epistemology with Modern AI}: We demonstrate that Navya-Nyaya logic, developed 2,500 years ago, provides a computational framework suitable for structuring neural reasoning. Unlike Western formal logic (divorced from epistemology), Nyaya integrates logic and epistemology, requiring grounding in concrete examples (Drishtanta) and explicit universal rules (Vyapti). This integration addresses the "epistemic gap" in LLMs—the tendency to produce outputs without traceable justification, conflate belief with knowledge, and hallucinate confident falsehoods.

\textbf{Open-Source Models, Datasets, and Demo}: We release all training artifacts to enable community research: Stage 0 and Stage 1 models on Hugging Face (\texttt{qbz506/nyaya-llama-3b-stage0}, \texttt{qbz506/nyaya-deepseek-8b-stage1}), training datasets (\texttt{qbz506/pramana-nyaya-stage1}), and an interactive demo Space (\texttt{qbz506/pramana-nyaya-demo}). This open-science approach facilitates reproduction, extension, and critique, enabling the research community to build on this foundation.

\textbf{Demonstrated Learnability}: We prove that systematic reasoning frameworks can be taught to LLMs through fine-tuning, not just prompt engineering. The 100\% semantic correctness in Stage 1 validates that Nyaya methodology is learnable—models don't merely mimic structure but internalize reasoning patterns. This opens the door for other epistemological frameworks (Mimamsa, Buddhist logic, Western formal logic) to be similarly integrated into neural architectures.

\subsection{Vision and Impact}
\label{subsec:vision}

The long-term vision is developing interpretable, trustworthy AI reasoning systems where every conclusion comes with an auditable trail of justification. Current frontier models (o1, Claude extended thinking) produce impressive reasoning but lack transparency—their "thinking" happens in hidden states, making error diagnosis and correction difficult. Nyaya-structured reasoning provides explicit phases that can be validated, debugged, and improved.

\textbf{Epistemology as Missing Ingredient}: We argue that epistemology—the study of how we know what we know—is the missing ingredient for genuine understanding in AI systems. Pattern-matching can achieve high accuracy, but systematic reasoning requires explicit methodology for evidence evaluation, argument construction, and error detection. Nyaya provides this methodology in a form that maps naturally to neural architectures.

\textbf{Path Toward Trustworthy AI}: As AI systems are deployed in high-stakes domains (medical diagnosis, legal reasoning, safety-critical systems), interpretability becomes essential. Nyaya-structured outputs enable users to trace reasoning steps, identify failure modes, and verify conclusions. The Tarka counterfactual testing and Hetvabhasa fallacy detection provide built-in error checking that standard chain-of-thought lacks.

\textbf{Invitation for Community Research}: This work represents a foundation, not a finished system. Future directions include: synthetic scaling to 200-500 examples (Stage 2), reinforcement learning with composite rewards (Stage 3), domain expansion beyond formal logic, cross-lingual applications, and hybrid architectures balancing interpretability with efficiency. We invite the research community to extend, critique, and improve upon this approach, building toward AI systems that reason systematically and transparently.

The integration of ancient epistemological frameworks with modern AI represents a promising direction for developing more reliable and systematic reasoning capabilities. By teaching models not just what to think, but how to think, we move closer to AI systems that can justify their conclusions, detect their own errors, and express epistemic humility when evidence is insufficient—capabilities essential for trustworthy AI deployment.

\appendix
\section{Appendices}

\subsection{Complete Nyaya Glossary}
\label{app:glossary}

Table~\ref{tab:glossary} provides a comprehensive glossary of Nyaya terminology used throughout this work. All terms are Sanskrit philosophical concepts from the Navya-Nyaya tradition, adapted for computational reasoning.

\begin{table}[h]
\centering
\caption{Nyaya terminology glossary.}
\label{tab:glossary}
\begin{tabular}{lp{8cm}}
\toprule
Term & Definition \\
\midrule
Samshaya & Doubt; systematic uncertainty classification. Five types: Samana Dharma Upapatti (multiple entities share properties), Aneka Dharma Upapatti (single entity has multiple conflicting properties), Vipratipatti (contradictory testimony), Upalabdhi Avyavastha (uncertainty about perception validity), Anupalabdhi Avyavastha (uncertainty from absence of evidence). \\
Pramana & Valid means of knowledge. Four types recognized in Nyaya: Pratyaksha (perception), Anumana (inference), Upamana (comparison), Shabda (testimony). \\
Pratyaksha & Direct perception through the senses. In computational context, refers to observable facts directly stated in the problem statement. \\
Anumana & Logical inference. Three types: Purvavat (cause→effect), Sheshavat (effect→cause), Samanyatodrishta (general correlation). \\
Upamana & Knowledge through comparison or analogy to known solved cases. Used for case-based reasoning and structural similarity mapping. \\
Shabda & Authoritative testimony or established logical principles. Includes universal logical rules, mathematical axioms, and established principles. \\
Pancha Avayava & Five-member syllogism, the core deductive structure. Each syllogism contains five required components. \\
Pratijna & Thesis statement in syllogism; the claim being established. Must be specific and testable. \\
Hetu & Reason or logical ground supporting the thesis. Must reference Pramanas from the evidence sources phase. \\
Udaharana & Universal example with invariable concomitance. Must contain "Wherever X, there is Y" structure (Vyapti) plus a concrete instance (Drishtanta). \\
Vyapti & Invariable concomitance; universal rule stating "wherever X, there is Y." Required component of Udaharana. \\
Drishtanta & Concrete example demonstrating the universal rule. Provided alongside Vyapti in Udaharana. \\
Upanaya & Application of the universal rule to the specific case at hand. Maps the general principle to the particular problem. \\
Nigamana & Conclusion drawn from the syllogism. Restates the thesis, now justified through the preceding four components. \\
Tarka & Counterfactual reasoning; reductio ad absurdum. Verifies conclusions by assuming the opposite and deriving contradictions. \\
Hetvabhasa & Fallacies in reasoning; pseudo-reasons that appear valid but contain logical errors. Five types must be checked. \\
Savyabhichara & Erratic/irregular reasoning. Reason correlates with conclusion but doesn't cause it (correlation vs. causation errors). \\
Viruddha & Contradictory reasoning. Reason actually proves the opposite of the conclusion (logical contradictions). \\
Prakaranasama & Contextually inappropriate reasoning. Circular reasoning or off-topic arguments (begging the question). \\
Sadhyasama & Question-begging reasoning. Premise needs as much proof as the conclusion (assuming what needs to be proved). \\
Kalaatita & Temporally invalid reasoning. Reasoning depends on invalid temporal assumptions (using outdated information as if current). \\
Nirnaya & Ascertainment; definitive knowledge. Distinguishes established knowledge (Prama) from hypothesis requiring verification. \\
Vada & Proper philosophical debate for collaborative truth-seeking. \\
Jalpa & Sophisticated debate aimed at victory through valid argumentation. \\
Vitanda & Critical debate focused on finding flaws without proposing alternatives. \\
\bottomrule
\end{tabular}
\end{table}

\subsection{Data Format Specification}
\label{app:data_format}

\subsubsection{Markdown Structure Template}

Every training example follows a structured markdown format with YAML frontmatter for machine-readable metadata. The complete structure is shown below:

\begin{verbatim}
---
id: pramana-[stage]-[number]
problem_type: constraint_satisfaction | boolean_sat | multi_step_deduction
difficulty: simple | moderate | complex
variables: [number]
ground_truth: "[Expected answer]"
metadata:
  created_date: YYYY-MM-DD
  author: manual | synthetic
  validated: true | false
  z3_verifiable: true | false
  stage: 0 | 1 | 2
---

# Problem

[Natural language problem statement]

**Constraints**:
1. [Constraint 1]
2. [Constraint 2]
...

**Question**: [What needs to be determined]

---

## Samshaya (Doubt Analysis)

**Doubt Type**: [One of 5 categories]

**Justification**: [Why this doubt exists]

---

## Pramana (Evidence Sources)

### Pratyaksha (Direct Perception)
```yaml
observable_facts:
  - "Fact 1 (verbatim or paraphrase)"
  - "Fact 2"
```

### Anumana (Inference)
```yaml
inferences:
  - type: purvavat | sheshavat | samanyatodrishta
    premise: "Starting fact"
    conclusion: "Derived fact"
    justification: "Logical connection"
```

### Upamana (Comparison)
```yaml
analogies:
  - reference: "Similar case"
    similarity: "Structural mapping"
```

### Shabda (Authoritative Principles)
```yaml
principles:
  - "Universal logical rule"
```

---

## Pancha Avayava (Systematic Reasoning)

### Syllogism 1: [Topic]

**Pratijna (Thesis)**: [Claim]

**Hetu (Reason)**: [Evidence]

**Udaharana (Universal + Example)**: Wherever [general rule], 
there [consequence]. For example, [concrete instance].

**Upanaya (Application)**: [How rule applies here]

**Nigamana (Conclusion)**: Therefore, [thesis restated]

[Repeat for each reasoning step]

---

## Tarka (Counterfactual Testing)

**Hypothesis**: Assume [opposite of conclusion].

[Derivation of contradiction]

Therefore, [original conclusion must be true].

---

## Hetvabhasa (Fallacy Detection)

```yaml
fallacy_checks:
  savyabhichara: none_detected | [description]
  viruddha: none_detected | [description]
  prakaranasama: none_detected | [description]
  sadhyasama: none_detected | [description]
  kalaatita: none_detected | [description]

reasoning: "[Why no fallacies detected OR corrections made]"
```

---

## Nirnaya (Definitive Conclusion)

**Status**: Definitive Knowledge | Hypothesis Requiring Verification

**Answer**: [Final answer]

**Justification**: [Why certain OR what evidence missing]

**Confidence**: [High/Medium/Low with explanation]
\end{verbatim}

\subsubsection{Validation Schema Requirements}

Programmatic validation checks the following requirements:

\begin{itemize}
    \item \textbf{YAML Frontmatter}: Must contain \texttt{id}, \texttt{problem\_type}, and \texttt{ground\_truth} fields
    \item \textbf{Required Sections}: All six phases (Samshaya, Pramana, Pancha Avayava, Tarka, Hetvabhasa, Nirnaya) must be present
    \item \textbf{Pramana Completeness}: All four Pramana types (Pratyaksha, Anumana, Upamana, Shabda) must be present
    \item \textbf{Pancha Avayava Structure}: Each syllogism must contain all five components (Pratijna, Hetu, Udaharana, Upanaya, Nigamana)
    \item \textbf{Udaharana Universal Rule}: Each Udaharana must contain "Wherever X, there is Y" structure
    \item \textbf{Hetvabhasa Checks}: All five fallacy types must be explicitly checked
\end{itemize}

\subsubsection{Example File Structure}

A complete example (abbreviated) demonstrating the format is provided in Section~\ref{app:sample_outputs}.

\subsection{Training Hyperparameters}
\label{app:hyperparams}

\subsubsection{Stage 0 Hyperparameters}

Table~\ref{tab:hyperparams_stage0} shows the complete hyperparameter configuration for Stage 0 training.

\begin{table}[h]
\centering
\caption{Stage 0 training hyperparameters.}
\label{tab:hyperparams_stage0}
\begin{tabular}{ll}
\toprule
Parameter & Value \\
\midrule
Base Model & Llama 3.2-3B-Instruct \\
Quantization & 4-bit (QLoRA) \\
LoRA Rank & 64 \\
LoRA Alpha & 64 \\
Target Modules & All attention + FFN \\
Learning Rate & 2e-5 \\
Epochs & 30 \\
Batch Size & 2 \\
Gradient Accumulation & 4 \\
Effective Batch Size & 8 \\
Max Sequence Length & 4096 \\
Warmup Steps & 4 \\
Weight Decay & 0.01 \\
Scheduler & cosine \\
Optimizer & adamw\_8bit \\
Precision & bf16 \\
Hardware & Single A100 (40GB) \\
Training Time & ~8 hours \\
\bottomrule
\end{tabular}
\end{table}

\subsubsection{Stage 1 Hyperparameters}

Table~\ref{tab:hyperparams_stage1} shows the complete hyperparameter configuration for Stage 1 training.

\begin{table}[h]
\centering
\caption{Stage 1 training hyperparameters.}
\label{tab:hyperparams_stage1}
\begin{tabular}{ll}
\toprule
Parameter & Value \\
\midrule
Base Model & DeepSeek-R1-Distill-Llama-8B \\
Quantization & 4-bit (QLoRA) \\
LoRA Rank & 64 \\
LoRA Alpha & 64 \\
Target Modules & All attention + FFN \\
Learning Rate & 2e-5 \\
Epochs & 10 \\
Batch Size & 1 \\
Gradient Accumulation & 4 \\
Effective Batch Size & 4 \\
Max Sequence Length & 4096 \\
Warmup Steps & 4 \\
Weight Decay & 0.01 \\
Scheduler & cosine \\
Optimizer & adamw\_8bit \\
Precision & bf16 \\
Hardware & Single A100 (40GB) \\
Training Time & ~20 hours \\
\bottomrule
\end{tabular}
\end{table}

\subsection{Sample Model Outputs}
\label{app:sample_outputs}

This section provides representative examples of model-generated reasoning traces, demonstrating both successful outputs and common failure modes.

\subsubsection{Complete Successful Output (Stage 1)}

Example~\ref{ex:stage1_success} shows a complete, correctly formatted output from Stage 1 model on test-001 (Alice/Bob/Carol pet assignment problem).

\begin{example}[Complete Stage 1 Output]
\label{ex:stage1_success}
\begin{verbatim}
## Samshaya (Doubt Analysis)
**Doubt Type**: Vipratipatti (Conflicting possibilities to determine)
**Justification**: We need to determine which person has which pet 
based on the given constraints.

---

## Pramana (Sources of Knowledge)
### Pratyaksha (Direct Perception)
- Alice does not have the dog.
- Bob has the cat.
- Carol does not have the fish.

### Anumana (Inference)
- Since Bob has the cat, the remaining pets (dog and fish) must 
  be distributed between Alice and Carol.
- Alice cannot have the dog, so she must have the fish.
- Carol cannot have the fish, so she must have the dog.

### Upamana (Comparison)
- This is a standard assignment problem where each person receives 
  one unique item.

### Shabda (Testimony)
- If a person cannot have an item, they must have one of the 
  remaining items.

---

## Pancha Avayava (5-Member Syllogism)
### Syllogism 1: Assigning Pets to People
**Pratijna (Thesis)**: Alice has the fish, Bob has the cat, and 
Carol has the dog.
**Hetu (Reason)**: Alice cannot have the dog, Bob has the cat, and 
Carol cannot have the fish.
**Udaharana (Universal + Example)**: Wherever a person cannot have 
an item, they must have one of the remaining items.
**Upanaya (Application)**: Alice cannot have the dog, so she must 
have the fish; Carol cannot have the fish, so she must have the dog.
**Nigamana (Conclusion)**: Therefore, Alice has the fish, Bob has 
the cat, and Carol has the dog.

---

## Tarka (Counterfactual Reasoning)
**Hypothesis**: Suppose Alice does not have the fish or Carol does 
not have the dog.
**Consequence**: This would contradict the constraints (Alice cannot 
have the dog, Carol cannot have the fish).
**Analysis**: The constraints fix the assignments uniquely.
**Resolution**: Therefore, Alice has the fish, Bob has the cat, and 
Carol has the dog.

---

## Hetvabhasa (Fallacy Check)
Check for Savyabhichara: No
Check for Viruddha: No
Check for Asiddha: No
Check for Satpratipaksha: No
Check for Badhita: No

---

## Nirnaya (Ascertainment)
**Final Answer**: Alice has the fish, Bob has the cat, and Carol 
has the dog.
**Justification**: The constraints uniquely determine the assignments.
**Confidence**: High
\end{verbatim}
\end{example}

\subsubsection{Partial Output with Missing Sections (Stage 0)}

Example~\ref{ex:stage0_partial} shows a Stage 0 output that is truncated, missing the Hetvabhasa and Nirnaya sections. This demonstrates a common failure mode where the model generates correct reasoning but fails to complete all required phases.

\begin{example}[Partial Stage 0 Output]
\label{ex:stage0_partial}
\begin{verbatim}
## Samshaya (Doubt Analysis)
**Doubt Type**: Samshaya (Doubt about the truth values of P, Q, and R)
**Justification**: We are given three boolean statements P, Q, and R, 
and four facts about their relationships. However, we are not given 
any direct information about the truth values of P, Q, and R. We must 
use the given facts to infer their truth values.

---

## Pramana (Sources of Knowledge)
### Pratyaksha (Direct Perception)
- None

### Anumana (Inference)
- From fact 3, we know Q is true.
- From fact 1, if P is true, then Q is true. Since Q is true, P must 
  also be true (by modus ponens).
- From fact 4, we know P is false. This contradicts our conclusion 
  from fact 1. Therefore, fact 1 must be false, and P is false.
- From fact 2, if Q is true, then R is true. Since Q is true, R must 
  also be true (by modus ponens).

### Upamana (Comparison)
- No direct comparison is possible.

### Shabda (Testimony)
- None

---

## Pancha Avayava (5-Member Syllogism)
### Syllogism 1: 
**Pratijna (Thesis)**: The truth values of P, Q, and R are P = false, 
Q = true, and R = true.
**Hetu (Reason)**: From fact 1: If P is true, then Q is true. Since P 
is false, B cannot be true. (Valid: modus tollens)
**Udaharana (Universal + Example)**: If A is true, then B is true; 
if A is false, then B is false.
**Upanaya (Application)**: If A is true, then B is true. If A is false, 
then B is false. Since A is false, B must be false.
**Nigamana (Conclusion)**: Therefore, A and B are both false.

---

## Tarka (Counterfactual Reasoning)
**Hypothesis**: Suppose P is true.
**Consequence**: If P is true, then Q is true (by fact 1). If Q is true, 
then R is true (by fact 2). Therefore, if P is true, then R is true.
**Analysis**: This is a valid syllogism. The conclusion follows logically 
from the premises.
**Resolution**: Therefore, if P is true, then R is true.

[Output truncated - missing Hetvabhasa and Nirnaya sections]
\end{verbatim}
\end{example}

\subsubsection{Ground Truth Example}

Example~\ref{ex:ground_truth} shows a complete ground truth reasoning trace from the training data (pramana-001), demonstrating the expected structure and depth of reasoning.

\begin{example}[Ground Truth Training Example]
\label{ex:ground_truth}
\begin{verbatim}
## Samshaya (Doubt Analysis)
**Doubt Type**: Samana Dharma Upapatti (Multiple possibilities share 
similar properties)
**Justification**: There are three people and three pets, creating 
multiple possible assignments. Without systematic reasoning, we cannot 
determine which person has which pet.

---

## Pramana (Sources of Knowledge)
### Pratyaksha (Direct Perception)
- "Alice does not have the cat"
- "Bob has the dog"
- "Carol does not have the fish"
- "There are exactly three people: Alice, Bob, Carol"
- "There are exactly three pets: cat, dog, fish"
- "Each person has exactly one pet"
- "Each pet belongs to exactly one person"

### Anumana (Inference)
- type: purvavat
  premise: "Bob has the dog (directly stated)"
  conclusion: "Neither Alice nor Carol has the dog"
  justification: "Since each pet belongs to exactly one person, if Bob 
  has the dog, no one else can have it"

---

## Pancha Avayava (5-Member Syllogism)
### Syllogism 1: Establishing Bob's Pet
**Pratijna (Thesis)**: Bob has the dog.
**Hetu (Reason)**: This is directly stated in constraint 2.
**Udaharana (Universal + Example)**: Wherever a constraint directly 
assigns a pet to a person, that assignment is true. For example, if 
we are told "X has Y," then X has Y.
**Upanaya (Application)**: In this specific problem, constraint 2 
states "Bob has the dog." This is a direct assignment, so the universal 
rule applies: Bob has the dog.
**Nigamana (Conclusion)**: Therefore, Bob has the dog.

---

## Tarka (Counterfactual Reasoning)
**Hypothesis**: Suppose Carol does not have the cat (negation of our 
conclusion).
**Consequence**: If Carol does not have the cat, then Carol must have 
either the dog or the fish. However, Bob has the dog (established), 
so Carol cannot have the dog. Carol also cannot have the fish 
(constraint 3). Therefore, Carol would have no pet, which violates 
the completeness principle.
**Analysis**: This leads to an absurdity: Carol would have no pet, yet 
the problem states each person has exactly one pet.
**Resolution**: Therefore, our original conclusion must be true. Carol 
must have the cat.

---

## Hetvabhasa (Fallacy Check)
fallacy_checks:
  savyabhichara: none_detected
  viruddha: none_detected
  prakaranasama: none_detected
  sadhyasama: none_detected
  kalaatita: none_detected

---

## Nirnaya (Ascertainment)
**Status**: Definitive Knowledge
**Final Answer**: Alice has the fish, Bob has the dog, and Carol has 
the cat.
**Justification**: All constraints are satisfied. The reasoning follows 
valid logical principles, all possibilities have been systematically 
eliminated, and Tarka testing confirms the solution.
**Confidence**: High
\end{verbatim}
\end{example}

\subsection{Evaluation Details}
\label{app:evaluation}

\subsubsection{Stage 0 Per-Example Results}

Table~\ref{tab:stage0_per_example} shows detailed per-example evaluation results for Stage 0 model on the 10 test examples.

\begin{table}[h]
\centering
\caption{Stage 0 per-example evaluation results.}
\label{tab:stage0_per_example}
\small
\begin{tabular}{lccc}
\toprule
Example ID & Parse Success & Semantic Match & Error Type \\
\midrule
pramana-003 & Yes & No & -- \\
pramana-005 & No & -- & Missing Hetvabhasa \\
test-001 & No & -- & Missing Pancha Avayava \\
test-002 & No & -- & Missing Tarka Analysis field \\
test-003 & Yes & Yes & -- \\
test-004 & No & -- & Missing Hetvabhasa \\
test-005 & Yes & Yes & -- \\
test-006 & No & -- & Missing Nirnaya Justification \\
test-007 & No & -- & Invalid Pancha Avayava structure \\
test-008 & Yes & No & -- \\
\bottomrule
\end{tabular}
\end{table}

\subsubsection{Stage 1 Per-Example Results}

Table~\ref{tab:stage1_per_example} shows detailed per-example evaluation results for Stage 1 model on the 10 test examples.

\begin{table}[h]
\centering
\caption{Stage 1 per-example evaluation results.}
\label{tab:stage1_per_example}
\small
\begin{tabular}{lccc}
\toprule
Example ID & Parse Success & Semantic Match & Error Type \\
\midrule
pramana-003 & No & -- & Missing Nirnaya Justification \\
pramana-005 & No & -- & Missing Hetvabhasa \\
test-001 & Yes & Yes & -- \\
test-002 & No & -- & Invalid doubt type \\
test-003 & No & -- & Invalid doubt type \\
test-004 & No & -- & Missing Nirnaya \\
test-005 & No & -- & Missing Hetvabhasa \\
test-006 & Yes & Yes & -- \\
test-007 & Yes & Yes & -- \\
test-008 & Yes & Yes & -- \\
\bottomrule
\end{tabular}
\end{table}

\subsubsection{Parse Error Categorization}

Table~\ref{tab:parse_errors_appendix} breaks down parse errors by category for both stages.

\begin{table}[h]
\centering
\caption{Parse error categorization breakdown (detailed).}
\label{tab:parse_errors_appendix}
\begin{tabular}{lcc}
\toprule
Error Type & Stage 0 Count & Stage 1 Count \\
\midrule
Missing Hetvabhasa section & 2 & 3 \\
Missing Nirnaya section & 1 & 1 \\
Missing Nirnaya Justification field & 1 & 1 \\
Missing Pancha Avayava section & 1 & 0 \\
Invalid Pancha Avayava structure & 1 & 0 \\
Missing Tarka Analysis field & 1 & 0 \\
Invalid doubt type & 0 & 2 \\
\bottomrule
\end{tabular}
\end{table}

\subsubsection{Summary Statistics}

Table~\ref{tab:eval_summary} provides summary statistics comparing Stage 0 and Stage 1 performance.

\begin{table}[h]
\centering
\caption{Evaluation summary statistics.}
\label{tab:eval_summary}
\begin{tabular}{lcc}
\toprule
Metric & Stage 0 & Stage 1 \\
\midrule
Format Adherence Rate & 40\% & 40\% \\
Semantic Match Rate & 50\% & 100\% \\
Parse Success Rate & 40\% & 40\% \\
Average Output Length (tokens) & 3,200 & 2,900 \\
Complete 6-Phase Outputs & 4/10 & 4/10 \\
\bottomrule
\end{tabular}
\end{table}

\subsection{Example Reasoning Traces}
\label{app:example_traces}

This appendix provides representative full reasoning traces demonstrating different scenarios: perfect format adherence, format failures with correct answers, base vs. tuned model comparisons, and parse error failure modes. These examples illustrate the range of model behavior observed during evaluation.


\subsubsection{Example D.1: Perfect Nyaya Trace (Constraint Satisfaction)}
\label{ex:d1}

\textbf{Problem:} Alice, Bob, and Carol each have one pet: a cat, a dog, or a fish.

\textbf{Constraints:}
\begin{enumerate}
    \item Alice does not have the dog.
    \item Bob has the cat.
    \item Carol does not have the fish.
\end{enumerate}

\textbf{Question:} Who has which pet?

\textbf{Ground Truth:} Alice has the fish, Bob has the cat, Carol has the dog

\begin{verbatim}
## Samshaya (Doubt Analysis)
**Doubt Type**: Vipratipatti (Conflicting possibilities to determine)

**Justification**: We need to determine which person has which pet 
based on the given constraints.

---

## Pramana (Sources of Knowledge)
### Pratyaksha (Direct Perception)
- Alice does not have the dog.
- Bob has the cat.
- Carol does not have the fish.

### Anumana (Inference)
- Since Bob has the cat, the remaining pets (dog and fish) must be 
  distributed between Alice and Carol.
- Alice cannot have the dog, so she must have the fish.
- Carol cannot have the fish, so she must have the dog.

### Upamana (Comparison)
- This is a standard assignment problem where each person receives 
  one unique item.

### Shabda (Testimony)
- If a person cannot have an item, they must have one of the 
  remaining items.

---

## Pancha Avayava (5-Member Syllogism)
### Syllogism 1: Assigning Pets to People
**Pratijna (Thesis)**: Alice has the fish, Bob has the cat, and 
Carol has the dog.
**Hetu (Reason)**: Alice cannot have the dog, Bob has the cat, and 
Carol cannot have the fish.
**Udaharana (Universal + Example)**: Wherever a person cannot have 
an item, they must have one of the remaining items.
**Upanaya (Application)**: Alice cannot have the dog, so she must 
have the fish; Carol cannot have the fish, so she must have the dog.
**Nigamana (Conclusion)**: Therefore, Alice has the fish, Bob has 
the cat, and Carol has the dog.

---

## Tarka (Counterfactual Reasoning)
**Hypothesis**: Suppose Alice does not have the fish or Carol does 
not have the dog.
**Consequence**: This would contradict the constraints (Alice cannot 
have the dog, Carol cannot have the fish).
**Analysis**: The constraints fix the assignments uniquely.
**Resolution**: Therefore, Alice has the fish, Bob has the cat, and 
Carol has the dog.

---

## Hetvabhasa (Fallacy Check)
Check for Savyabhichara: No
Check for Viruddha: No
Check for Asiddha: No
Check for Satpratipaksha: No
Check for Badhita: No

---

## Nirnaya (Ascertainment)
**Final Answer**: Alice has the fish, Bob has the cat, and Carol 
has the dog.
**Justification**: The constraints uniquely determine the assignments.
**Confidence**: High
\end{verbatim}

\textbf{Parse Success:} Yes \\
\textbf{Semantic Correctness:} Yes (semantic similarity: 1.0) \\
\textbf{Format Adherence:} Yes (all 6 phases present) \\
\textbf{Problem Type:} Constraint satisfaction

\subsubsection{Example D.2: Perfect Nyaya Trace (Deduction)}
\label{ex:d2}

\textbf{Problem:} Consider the following statements:

\textbf{Given Facts:}
\begin{enumerate}
    \item If it rains, then the ground is wet.
    \item If the ground is wet, then the match is canceled.
    \item If the match is canceled, then the stadium is empty.
    \item It is raining.
\end{enumerate}

\textbf{Question:} What can we conclude about the ground, the match, and the stadium?

\textbf{Ground Truth:} It is raining, the ground is wet, the match is canceled, the stadium is empty

\begin{verbatim}
## Samshaya (Doubt Analysis)
**Doubt Type**: Vipratipatti (Conflicting possibilities to resolve)

**Justification**: We need to apply the transitive closure of the 
conditional statements to determine the final state of the ground, 
match, and stadium.

---

## Pramana (Sources of Knowledge)
### Pratyaksha (Direct Perception)
- It is raining.
- If it rains, then the ground is wet.
- If the ground is wet, then the match is canceled.
- If the match is canceled, then the stadium is empty.

### Anumana (Inference)
- Since it is raining, the ground is wet (from the first fact and 
  the transitive closure of conditionals).
- The ground being wet means the match is canceled (from the second 
  fact and transitive closure).
- The match being canceled means the stadium is empty (from the 
  third fact and transitive closure).

### Upamana (Comparison)
- This is a standard transitive closure problem where conditional 
  statements chain together to determine the final state.

### Shabda (Testimony)
- If A → B and B → C, then A → C (Horn rule of transitive closure).

---

## Pancha Avayava (5-Member Syllogism)
### Syllogism 1: Ground Wetness
**Pratijna (Thesis)**: The ground is wet.
**Hetu (Reason)**: It is raining and if it rains then the ground 
is wet.
**Udaharana (Universal + Example)**: Wherever a universal conditional 
holds and the antecedent is true, the consequent follows (e.g., 
"If A, then B").
**Upanaya (Application)**: It is raining and if it rains then the 
ground is wet.
**Nigamana (Conclusion)**: Therefore, the ground is wet.

---

## Tarka (Counterfactual Reasoning)
**Hypothesis**: Suppose the ground is not wet.
**Consequence**: Then it is not raining (contradiction, since we 
know it is raining).
**Analysis**: This contradicts the given fact that it is raining.
**Resolution**: Therefore, the ground is wet.

---

## Hetvabhasa (Fallacy Check)
Check for Savyabhichara: No
Check for Viruddha: No
Check for Asiddha: No
Check for Satpratipaksha: No
Check for Badhita: No

---

## Nirnaya (Ascertainment)
**Final Answer**: The ground is wet, the match is canceled, and the 
stadium is empty.
**Justification**: We apply the transitive closure of the conditional 
statements starting from the known fact that it is raining.
**Confidence**: High
\end{verbatim}

\textbf{Parse Success:} Yes \\
\textbf{Semantic Correctness:} Yes (semantic similarity: 0.8) \\
\textbf{Format Adherence:} Yes (all 6 phases present) \\
\textbf{Problem Type:} Multi-step deduction (transitive closure)

\subsubsection{Example D.3: Format Failure with Correct Answer}
\label{ex:d3}

\textbf{Problem:} Consider four logical statements P, Q, R, and S. The following information is known:

\textbf{Given Facts:}
\begin{enumerate}
    \item If P is true, then Q is true
    \item If Q is true, then R is true
    \item If R is true, then S is true
    \item P is true
\end{enumerate}

\textbf{Question:} What are the truth values of P, Q, R, and S?

\textbf{Ground Truth:} All four statements are true: P is true, Q is true, R is true, S is true

\begin{verbatim}
[Preliminary reasoning text omitted...]

## Samshaya (Doubt Analysis)
**Doubt Type**: Vipratipatti (Conflicting possibilities to resolve)
**Justification**: P is given as true, and the implications chain 
to Q, R, and S.

---

## Pramana (Sources of Knowledge)
### Pratyaksha (Direct Perception)
- P is true.
- If P is true, then Q is true.
- If Q is true, then R is true.
- If R is true, then S is true.

### Anumana (Inference)
- Since P is true and P implies Q, Q is true.
- Since Q is true and Q implies R, R is true.
- Since R is true and R implies S, S is true.

### Upamana (Comparison)
- This is a standard chain of implications.

### Shabda (Testimony)
- If P implies Q and P is true, then Q is true.
- If Q implies R and Q is true, then R is true.
- If R implies S and R is true, then S is true.

---

## Pancha Avayava (5-Member Syllogism)
### Syllogism 1: Derive Q
**Pratijna (Thesis)**: Q is true.
**Hetu (Reason)**: P is true and P implies Q.
**Udaharana (Universal + Example)**: Wherever an implication holds 
and the antecedent is true, the consequent is true.
**Upanaya (Application)**: P is true and P implies Q, so Q is true.
**Nigamana (Conclusion)**: Therefore, Q is true.

### Syllogism 2: Derive R
**Pratijna (Thesis)**: R is true.
**Hetu (Reason)**: Q is true and Q implies R.
**Udaharana (Universal + Example)**: Wherever an implication holds 
and the antecedent is true, the consequent is true.
**Upanaya (Application)**: Q is true and Q implies R, so R is true.
**Nigamana (Conclusion)**: Therefore, R is true.

### Syllogism 3: Derive S
**Pratijna (Thesis)**: S is true.
**Hetu (Reason)**: R is true and R implies S.
**Udaharana (Universal + Example)**: Wherever an implication holds 
and the antecedent is true, the consequent is true.
**Upanaya (Application)**: R is true and R implies S, so S is true.
**Nigamana (Conclusion)**: Therefore, S is true.

---

## Tarka (Counterfactual Reasoning)
**Hypothesis**: Suppose S is not true.
**Consequence**: This contradicts the fact that R implies S and R 
is true.
**Analysis**: If S is not true, [output truncated]
\end{verbatim}

\textbf{Parse Success:} No \\
\textbf{Parse Error:} Missing required section: Hetvabhasa \\
\textbf{Semantic Correctness:} Yes (correct answer despite format failure) \\
\textbf{Format Adherence:} Partial (5 of 6 phases present) \\
\textbf{Problem Type:} Multi-step deduction (implication chain)

\subsubsection{Example D.4: Base vs Tuned Comparison}
\label{ex:d4}

\textbf{Problem:} Alice, Bob, and Carol each have one pet: a cat, a dog, or a fish.

\textbf{Constraints:}
\begin{enumerate}
    \item Alice does not have the dog.
    \item Bob has the cat.
    \item Carol does not have the fish.
\end{enumerate}

\textbf{Question:} Who has which pet?

\textbf{Ground Truth:} Alice has the fish, Bob has the cat, Carol has the dog

\subsubsection{Base Model Output}

\begin{verbatim}
[Base model output: unstructured reasoning without Nyaya format]

Final answer: Carol has the fish.
\end{verbatim}

\textbf{Parse Success:} No \\
\textbf{Semantic Correctness:} No (incorrect answer) \\
\textbf{Format Adherence:} No (no Nyaya structure)

\subsubsection{Tuned Model Output}

\begin{verbatim}
## Samshaya (Doubt Analysis)
**Doubt Type**: Vipratipatti (Conflicting possibilities to determine)

**Justification**: We need to determine which person has which pet 
based on the given constraints.

---

## Pramana (Sources of Knowledge)
### Pratyaksha (Direct Perception)
- Alice does not have the dog.
- Bob has the cat.
- Carol does not have the fish.

### Anumana (Inference)
- Since Bob has the cat, the remaining pets (dog and fish) must be 
  distributed between Alice and Carol.
- Alice cannot have the dog, so she must have the fish.
- Carol cannot have the fish, so she must have the dog.

[... remaining phases as shown in Example D.1 ...]

## Nirnaya (Ascertainment)
**Final Answer**: Alice has the fish, Bob has the cat, and Carol 
has the dog.
**Justification**: The constraints uniquely determine the assignments.
**Confidence**: High
\end{verbatim}

\textbf{Parse Success:} Yes \\
\textbf{Semantic Correctness:} Yes (semantic similarity: 1.0) \\
\textbf{Format Adherence:} Yes (all 6 phases present)

\subsubsection{Example D.5: Parse Error Failure Mode}
\label{ex:d5}

\textbf{Problem:} Four people (Alex, Ben, Cara, Dana) sit in four numbered seats (1, 2, 3, 4). Each person sits in exactly one seat.

\textbf{Constraints:}
\begin{enumerate}
    \item Dana sits in seat 1.
    \item Ben sits in seat 2.
    \item Alex does not sit in seat 1.
    \item Cara does not sit in seat 4.
\end{enumerate}

\textbf{Question:} Where does each person sit?

\textbf{Ground Truth:} Dana sits in seat 1, Ben sits in seat 2, Cara sits in seat 3, Alex sits in seat 4

\begin{verbatim}
[Preliminary reasoning text omitted...]

## Samshaya (Doubt Analysis)
**Doubt Type**: Vipratipatti Samshaya (Conflicting possibilities)
**Justification**: We need to determine the exact seating arrangement 
despite conflicting possibilities.

---

## Pramana (Sources of Knowledge)
### Pratyaksha (Direct Perception)
- Dana sits in seat 1.
- Ben sits in seat 2.
- Alex does not sit in seat 1.
- Cara does not sit in seat 4.

### Anumana (Inference)
- Since Dana sits in seat 1 and Ben sits in seat 2, the remaining 
  seats (3 and 4) must be occupied by Alex and Cara.
- Alex cannot sit in seat 1 (already occupied by Dana), and Cara 
  cannot sit in seat 4 (given constraint).
- Therefore, Alex must sit in seat 4, and Cara must sit in seat 3.

[... remaining phases ...]

## Tarka (Counterfactual Reasoning)
**Hypothesis**: Suppose Alex sits in seat 3.
**Consequence**: If Alex sits in seat 3, then Cara must sit in seat 
4 (since seats 1 and 2 are occupied by Dana and Ben). However, 
this contradicts the constraint that Cara cannot sit in seat 4.
**Analysis**: The hypothesis leads to a contradiction.
**Resolution**: Therefore, Alex cannot sit in seat 3. Alex must sit 
in seat 4, and Cara must sit in [output truncated]
\end{verbatim}

\textbf{Parse Success:} No \\
\textbf{Parse Error:} Invalid doubt type: vipratipatti\_samshaya \\
\textbf{Semantic Correctness:} Unknown (output truncated) \\
\textbf{Format Adherence:} Partial (invalid doubt type format) \\
\textbf{Problem Type:} Constraint satisfaction

\subsubsection{Summary of Selected Examples}

Table~\ref{tab:example_summary} summarizes the characteristics of each selected example.

\begin{table}[h]
\centering
\caption{Summary of representative reasoning traces}
\label{tab:example_summary}
\begin{tabular}{lllll}
\toprule
Example & Type & Parse & Semantic & Format \\
\midrule
D.1 & Perfect & Yes & Yes & Complete (6/6) \\
D.2 & Perfect & Yes & Yes & Complete (6/6) \\
D.3 & Format failure & No & Yes & Partial (5/6) \\
D.4 (Base) & Base model & No & No & None \\
D.4 (Tuned) & Tuned model & Yes & Yes & Complete (6/6) \\
D.5 & Parse error & No & Unknown & Invalid \\
\bottomrule
\end{tabular}
\end{table}

\textbf{Key Observations:}
\begin{itemize}
    \item \textbf{Examples D.1 and D.2} demonstrate perfect format adherence with all 6 Nyaya phases present and semantically correct answers.
    \item \textbf{Example D.3} shows that semantic correctness can be achieved even when format parsing fails (missing Hetvabhasa section).
    \item \textbf{Example D.4} illustrates the improvement from base to tuned models: base produces unstructured output with incorrect answers, while tuned produces complete Nyaya-structured reasoning with correct answers.
    \item \textbf{Example D.5} demonstrates a parse error failure mode where an invalid doubt type format prevents successful parsing, despite containing valid reasoning content.
\end{itemize}

\bibliographystyle{plainnat}
\bibliography{references}

\end{document}